\documentclass[year=22]{fmcad}

\usepackage{alltt}
\usepackage{caption}
\usepackage{paralist}
\usepackage{nicefrac}
\usepackage{booktabs}
\usepackage{algorithm}
\usepackage{algpseudocode}
\usepackage{xspace}
\usepackage{amsmath}
\usepackage{svg}
\usepackage{marvosym}
\usepackage{wasysym}
\usepackage{verbatim}
\usepackage{subfigure}
\usepackage{hyperref}
\usepackage{multirow}
\usepackage{cite}
\usepackage[capitalize]{cleveref}
\usepackage[per-mode=symbol,detect-all]{siunitx}
\usepackage{graphics}
\usepackage{amssymb}
\usepackage{wrapfig}
\usepackage{enumitem}
\usepackage{placeins}

\usepackage{tikz,ifthen,pgfplots}

\newcommand{\mysubsection}[1]{\medskip\noindent\textbf{#1}}

\usetikzlibrary{arrows,trees,backgrounds,automata,shapes,decorations,plotmarks,fit,calc,positioning,shadows,chains}

\definecolor{color0}{RGB}{228,87,46}
\definecolor{color1}{RGB}{23,190,187}
\definecolor{color2}{RGB}{255,201,20}
\definecolor{color3}{RGB}{46,40,42}
\definecolor{color4}{RGB}{118,176,65}

\definecolor{color0}{RGB}{250,121,33}
\definecolor{color1}{RGB}{254,153,32}
\definecolor{color2}{RGB}{185,164,76}
\definecolor{color3}{RGB}{86,110,61}
\definecolor{color4}{RGB}{12,71,103}

\definecolor{color0}{RGB}{228,253,225}
\definecolor{color1}{RGB}{138,203,136}
\definecolor{color2}{RGB}{100,131,129}
\definecolor{color3}{RGB}{87,87,97}
\definecolor{color4}{RGB}{255,191,70}

\definecolor{color0}{RGB}{226,59,62}
\definecolor{color1}{RGB}{243,114,44}
\definecolor{color2}{RGB}{248,150,30}
\definecolor{color3}{RGB}{249,199,79}
\definecolor{color4}{RGB}{126,179,86}
\definecolor{color5}{RGB}{67,170,139}
\definecolor{color6}{RGB}{39,125,161}
\definecolor{color7}{RGB}{21,49,60}
\definecolor{color8}{RGB}{180,215,228}

\tikzstyle{every pin edge}=[<-,shorten <=1pt]
\tikzstyle{neuron}=[circle,fill=black!25,minimum size=17pt,inner sep=0pt]
\tikzstyle{input neuron}=[neuron, fill=green!50]
\tikzstyle{output neuron}=[neuron, fill=red!50]
\tikzstyle{hidden neuron}=[neuron, fill=blue!50]
\tikzstyle{annot} = [text width=4em, text centered]
\usetikzlibrary{calc}

\newcommand{\vecx}{\boldsymbol{x}}

\newcommand{\relu}{\text{ReLU}\xspace{}}
\newcommand{\frelu}[1]{\relu(#1)}
\newcommand{\fsig}[1]{\sigma(#1)}
\newcommand{\fmax}[1]{\max(#1)}
\newcommand{\fmin}[1]{\min(#1)}

\newcommand{\inputDomain}{\mathcal{D}_i}
\newcommand{\outputDomain}{\mathcal{D}_o}

\definecolor{nnedgecolor}{RGB}{90,90,90}
\tikzstyle{every pin edge}=[<-,shorten <=1pt]
\tikzstyle{every path}=[draw=color7!50]
\tikzstyle{neuron}=[circle,fill=black!25,minimum size=17pt,inner sep=0pt]
\tikzstyle{input neuron}=[neuron, fill=color4]
\tikzstyle{output neuron}=[neuron, fill=color0]
\tikzstyle{hidden neuron}=[neuron, fill=color6!80]
\tikzstyle{annot} = [text width=4em, text centered]
\tikzstyle{nnedge} = [-{stealth},shorten >=0.1cm, shorten <=0.05cm,line width=0.8pt,nnedgecolor]
\usetikzlibrary{calc}

\newtheorem{definition}{\textbf{Definition}}

\begin{document}

\title{On Optimizing Back-Substitution Methods for Neural Network Verification}

\author{
  \IEEEauthorblockN{
    Tom Zelazny\IEEEauthorrefmark{1},
    Haoze Wu\IEEEauthorrefmark{2},
    Clark Barrett\IEEEauthorrefmark{2},
    and
    Guy Katz\IEEEauthorrefmark{1}
  }
  \IEEEauthorblockA{
    \IEEEauthorrefmark{1}The Hebrew University of Jerusalem, Jerusalem, Israel
    \IEEEauthorrefmark{2}Stanford University, Stanford, California
  }
  \IEEEauthorblockA{
    \IEEEauthorrefmark{1}$ \lbrace tomz, g.katz \rbrace $@mail.huji.ac.il
    \IEEEauthorrefmark{2}$ \lbrace haozewu, barrett \rbrace $@cs.stanford.edu
  }
}

\maketitle

\begin{abstract}
  With the increasing application of deep learning in mission-critical
  systems, there is a growing need to obtain formal guarantees about
  the behaviors of neural networks. Indeed, many approaches for
  verifying neural networks have been recently proposed, but these
  generally struggle with limited scalability or insufficient
  accuracy. A key component in many state-of-the-art verification
  schemes is computing lower and upper bounds on the values that
  neurons in the network can obtain for a specific input domain ---
  and the tighter these bounds, the more likely the verification is to
  succeed. Many common algorithms for computing these bounds are
  variations of the symbolic-bound propagation method; and among
  these, approaches that utilize a process called back-substitution
  are particularly successful. In this paper, we present an approach
  for making back-substitution produce tighter bounds. To achieve
  this, we formulate and then minimize the imprecision errors incurred
  during back-substitution.
  %We explore two variations of the approach,
  %each presenting a different tightness/run-time trade-off.
  Our technique is general, in the sense that it can be integrated
  into numerous existing symbolic-bound propagation techniques, with
  only minor modifications. We implement our approach as a
  proof-of-concept tool, and present favorable results compared
  to state-of-the-art verifiers that perform back-substitution.
\end{abstract}

\section{Introduction}\label{sec:introduction}

Deep neural networks (DNNs) are dramatically changing the way modern
software is written. In many domains, such as image
recognition~\cite{SiZi14}, game
playing~\cite{SiHuMaGuSiVaScAnPaLaDi16}, protein folding~\cite{Al19}
and autonomous vehicle control~\cite{BoDeDwFiFlGoJaMoMuZhZhZhZi16,
  JuLoBrOwKo16}, state-of-the-art solutions involve deep neural
networks --- which are artifacts learned automatically from a finite
set of examples, and which often outperform carefully handcrafted
software.

Along with their impressive success, DNNs present a significant new
challenge when it comes to quality assurance. Whereas many best
practices exist for writing, testing, verifying and maintaining
hand-crafted code, DNNs are automatically generated, and are mostly
opaque to humans~\cite{Gu17, FoBeCu16}. Consequently, it is difficult
for human engineers to reason about them and ensure their correctness
and safety --- as most existing approaches are ill-suited for this
task. This challenge is becoming a significant concern, with various
faults being observed in modern DNNs~\cite{AmOlStChScMa16}. One
notable example is that of \emph{adversarial perturbations} --- small
perturbation that, when added to inputs that are correctly classified
by the DNN, result in severe errors~\cite{SzZaSuBrErGoFe13,
  EyEvFeLiRaXiPrKoSo18}. This issue, and others, call into question
the safety, security and interpretability of DNNs, and could hinder
their adoption by various stakeholders.

In order to mitigate this challenge, the formal methods community has
taken up interest in DNN verification. In the past few years, a
plethora of approaches have been proposed for tackling the \emph{DNN
  verification problem}, in which we are given a DNN and a condition
abouts its inputs and outputs; and seek to either find an input
assignment to the DNN that satisfies this condition, or prove that it
is not satisfiable~\cite{KaBaDiJuKo21, HuKwWaWu17, WaPeWhYaJa18,
  GeMiDrTsChVe18, PuTa10, ZhShGuGuLeNa20, AvBlChHeKoPr19,
  BaShShMeSa19, AkKeLoPi19, DrFrGhKiRaVaSe19, KoKoKiAtAs20, JiTiZhWeZh22}.
 %The condition often represents the negation of the desired property of the network, and so proving that it cannot be met implies the correctness of the network. 
The
usefulness of DNN verification has been demonstrated in several
settings and domains~\cite{KaBaDiJuKo21, GeMiDrTsChVe18, SuKhSh19,
  HuKwWaWu17}, but most existing approaches still struggle with
various limitations, specifically relating to scalability.

A key technical challenge in verifying neural networks is to reason
about \emph{activation functions}, which are non-linear (e.g.,
piece-wise linear) transformations applied to the output of each layer
in the neural network. Precisely reasoning about such non-linear
behaviors requires a case-by-case analysis of the activation phase of
each activation function, which quickly becomes infeasible as the
number of non-linear activations increases. Instead, before performing
such a search procedure, state-of-the-art solvers typically first
consider linear abstractions of activation functions, and use these
abstractions to over-approximate the values that the activation
functions can take in the neural network. Often, these
over-approximations significantly curtail the search space that later
needs to be explored, and expedite the verification procedure as a
whole.

A key operation that is repeatedly invoked in this computation of
over-approximations is called
\emph{back-substitution}~\cite{GaGePuVe19}, where the goal is to
compute, for each neuron in the DNN, lower and upper bounds on the
values it can take with respect to the input region of interest. This
is done by first expressing the lower and upper bounds of a neuron
symbolically as a function of neurons from previous layers, and then
concretizing these symbolic bounds with the known bounds of neurons in
those previous layers. Such a technique is essential in
state-of-the-art solvers (e.g.,
\cite{GaGePuVe19,XuZhWaWaJaLiHs20,KaHuIbJuLaLiShThWuZeDiKoBa19}) and
is often able to obtain sufficiently tight bounds for proving the
properties with respect to small input regions. However, it tends to
significantly lose precision when the input region (i.e., perturbation
radius) grows, preventing one from efficiently verifying more
challenging problems.

In this work, we seek to %contribute to this ongoing effort, by
improve the precision and scalability of DNN verification
techniques, by reducing the over-approximation error in the back-substitution process.
%We focus here on a specific technique for DNN
%verification, called \emph{back-substitution}~\cite{GaGePuVe19},
%which is among the most successful techniques today. In
%back-substitution, our goal is to compute, for each neuron in the DNN,
%lower and upper bounds on the values it can obtain in the input region
%of interest. Obtaining tight bounds for intermediate neurons allows in
%turn to obtain tight bounds for neurons in the network's output layer,
%and this is often sufficient for proving that the DNN is
%safe. Unfortunately, computing tight bounds is difficult, and so one
%often settles for approximate bounds~\cite{GaGePuVe19, Eh17}. In
%back-substitution, we seek to express the dependencies between neurons
%as a function of neurons from previous layers (e.g., the input layer);
%and then, by observing the known bounds for neurons in those previous
%layers, deduce bounds for the neurons in the current layer. The
%success of back-substitution techniques is due primarily to the
%effective balance between accuracy and speed that they
%provide~\cite{BaLiJo21}
Our key insight is that, as part of the symbolic-bound propagation, one
can measure the error accumulated by the over-approximations used in
back-substitution. Often, the currently computed bound can then be
significantly improved by ``pushing'' it towards the
true function, in a way that maintains its validity.
For example, suppose that we upper-bound a function $f$ with
a function $g$, i.e.  $\forall x.\ g(x)\geq f(x)$. If we discover that
the minimal approximation error is $5$, i.e. $\min_x\{g(x)-f(x)\}=5$,
then $g(x)-5$ can be used as a better upper bound for $f$ than
the original $g$. By integrating this simple principle into the
back-substitution process, we show that we can obtain much tighter
bounds, which eventually translates to the ability to verify more
difficult properties.

We propose here a verification approach, called \emph{DeepMIP}, that
uses symbolic-bound tightening enhanced with our error-optimization
method. At each iteration of the back-substitution, DeepMIP invokes an
external MIP solver~\cite{Gurobi} to compute bounds on the error of
the current approximation, and then uses these bounds to
improve that approximation. As we show, this leads to an improved
ability to solve verification benchmarks when compared to
state-of-the-art, symbolic-bound tightening techniques. We discuss the
different advantages of the approach, as well as the extra overhead
that it incurs, and various enhancements that could be used to
expedite it further.

The rest of the paper is organized as follows. We begin by presenting
the necessary background on DNNs, DNN verification, and on
symbolic-bound propagation in Sec.~\ref{sec:background}. Next, in
Sec.~\ref{sec:errorsinBS} we show how one can express the
approximation error incurred as part of the back-substitution process.
In Sec.~\ref{sec:DeepMIP} we present the DeepMIP algorithm, followed
by its evaluation in Sec.~\ref{sec:evaluation}. Related work is
discussed in Sec.~\ref{sec:relatedWork}, and we conclude in
Sec.~\ref{sec:conclusion}.

\section{Background}
\label{sec:background}

\mysubsection{Neural networks.} 
%For simplicity, we focus on fully-connected feed-forward neural networks that use the $\relu$ activation function,
%defined as $\relu(x)=\max\{0,x\}$. 
A fully-connected feed-forward neural network with $k+1$ layers is a
function $N:\mathbb{R}^{m}\rightarrow\mathbb{R}^n$. Given an input
$\vecx\in\mathbb{R}^m$, we use $N_i(\vecx)$ to denote the values of neurons in
the $i^{th}$ layer ($0\leq i \leq k$). The output of the neural
network $N(\vecx)$ is defined as $N_k(\vecx)$, which we refer to as the output
layer. More concretely, for $1\leq i\leq k$,
\[
  N_i(\vecx) = \sigma(W^{i-1}N_{i-1}(\vecx) + b^{i-1})
\]
where $W^{i-1}$ is a \emph{weight matrix}, $b^{i-1}$ is a \emph{bias vector}, $\sigma$ is an activation function (in this paper, we focus on the $\relu$ activation function, defined as $\relu(x)=\max\{0,x\}$ and use $\sigma$ and $\relu$ interchangeably unless otherwise specified) and $N_{0}\left(\vecx\right) = \vecx$. We refer to $N_0$ as the input layer. Typically, non-linear activations are not applied to the output layer. Thus, when $i = k$, we let $\sigma$ be the identity function. We note that our techniques are general, and apply to other activation functions (MaxPool, LeakyReLU) and architectures (e.g., convolutional, residual).
 
\mysubsection{Neural network verification.}  The \emph{neural network
  verification problem}~\cite{PuTa10, KaBaDiJuKo21} is defined as
follows: given an input domain $\inputDomain\subseteq \mathbb{R}^{m}$
and an output domain domain $\outputDomain\subseteq \mathbb{R}^n$,
the goal is to determine whether
$\forall \vecx \in \inputDomain, N(\vecx)\in\outputDomain$.  If the
answer is affirmative, we say that the verification property pair
$\langle \inputDomain, \outputDomain \rangle$ holds. In this paper, we assume that the neural network has a
single output neuron and that the verification problem can be reduced to the problem of finding the minimum and/or maximum values for that single output neuron:
\begin{equation}
  \label{eq:verificationProblem}
  \min_{\vecx\in\mathcal{D}_i}(N (\vecx))
  \qquad
  \max_{\vecx\in\mathcal{D}_i}(N (\vecx))
\end{equation}
% It has been shown that, in practice, a wide range of verification queries (including adversarial robustness) can be transformed to meet these two assumptions in a straightforward way~\cite{ElGoKa20}.
% and then checking whether these values intersect $\outputDomain$
% (here, $N(\vecx)$ is the scalar value of the single output neuron). 
For example, if $\outputDomain$ is the interval $[-2,7]$ and we discover
that $\min_{x\in\mathcal{D}_i}(N (x)) = 1$ and
$\max_{x\in\mathcal{D}_i}(N (x)) = 3$, then we are guaranteed that the
property holds.  
We will focus on solving just the
maximization problem, although the method that we present next can just as readily be applied towards the minimization problem.

A straightforward way to solve the optimization
problem in Eq.~\ref{eq:verificationProblem} is to encode the neural network
as a mixed integer programming (MIP) instance~\cite{TjXiTe17,
  BaIoLaVyNoCr16, KaBaDiJuKo21}, and then solve the problem using a MIP solver, which often employs a
branch-and-bound procedure. % an off-the-shelf solver
%to dispatch it. %The performance of this methods depends heavily on the encoding used for the activation functions~\cite{TjXiTe17}, and on the property being verified. %Presently, approaches that rely solely on MIP encodings
While this approach has proven effective at verifying small DNNs, it
faces a scalability barrier when it comes to larger
networks. %Still, MIP solvers have been effectively combined with other verification methods, and integrated into many modern DNN verifiers~\cite{KaHuIbJuLaLiShThWuZeDiKoBa19, MuMaSiPuVe22, XuZhWaWaJaLiHs20}.
Therefore, before invoking the branch-and-bound procedure, existing
solvers typically first seek to prove the property with
abstraction-based techniques (symbolic-bound propagation), which have
more tractable runtime.

% instances however due to the nature of the problem it does not scale to larger networks. Still, MIP solvers have shown great efficiency at solving smaller instances and are used by many state-of-the-art frameworks to find optimal bounds for neurons at early layers of networks.  As errors often accumulate stronger (optimal) bounds even only for the early layer, have a very positive effect on the success of many verification methods.

\mysubsection{Symbolic-bound propagation.}  Symbolic-bound
propagation~\cite{WaPeWhYaJa18, GeMiDrTsChVe18} is a method of
obtaining bounds on the concrete values a neuron may obtain. When
applied to a network's output neuron, it enables us to obtain an
approximate solution to the optimization problems from
Eq.~\ref{eq:verificationProblem}, which may be sufficient to determine
that the property holds. For example, continuing the example from
before, if we are unable to exactly compute that
$\max_{x\in\mathcal{D}_i}(N (x)) = 3$ but can determine that
$\max_{x\in\mathcal{D}_i}(N (x)) < 5$, this is enough for concluding
that the property in question holds.  The idea underlying
symbolic-bound propagation is to start from the bounds for the input
layer provided in $\inputDomain$, and then propagate them,
layer-by-layer, up to the output layer. It has been observed that
while affine transformations allow us to precisely propagate bounds
from a layer to its successor, activation functions introduce
inaccuracies~\cite{GaGePuVe19}.

\begin{figure}[ht]
  \centering
  \scalebox{0.8}{
  \def\WSsep{2.0cm}
  \def\relusep{1.5cm}
  \begin{tikzpicture}[shorten >=1pt,->,draw=black!50, node distance=\layersep,font=\footnotesize]
    
    \node[input neuron] (I-1) at (0,0) {$x^0_0$};
    \node[input neuron] (I-2) at (0,-2) {$x^0_1$};
    \node[input neuron] (I-3) at (0,-4) {$x^0_2$};

    \node[hidden neuron] (H-1) at (\WSsep + 0.3cm,0) {$x^1_0$};
    \node[hidden neuron] (H-2) at (\WSsep + 0.3cm,-2) {$x^1_1$};
    \node[hidden neuron] (H-3) at (\WSsep + 0.3cm,-4) {$x^1_2$};
    
    \node[hidden neuron] (H-4) at (\WSsep + \relusep + 0.3cm,0) {$y^1_0$};
    \node[hidden neuron] (H-5) at (\WSsep + \relusep + 0.3cm,-2) {$y^1_1$};
    \node[hidden neuron] (H-6) at (\WSsep + \relusep + 0.3cm,-4) {$y^1_2$};
    
    \node[hidden neuron] (H-7) at (\relusep + 2*\WSsep + 0.3cm,0) {$x^2_0$};
    \node[hidden neuron] (H-8) at (\relusep + 2*\WSsep + 0.3cm,-2) {$x^2_1$};
    \node[hidden neuron] (H-9) at (\relusep + 2*\WSsep + 0.3cm,-4) {$x^2_2$};

    \node[hidden neuron] (H-10) at (2*\relusep + 2*\WSsep + 0.3cm,0) {$y^2_0$};
    \node[hidden neuron] (H-11) at (2*\relusep + 2*\WSsep + 0.3cm,-2) {$y^2_1$};
    \node[hidden neuron] (H-12) at (2*\relusep + 2*\WSsep + 0.3cm,-4) {$y^2_2$};

    \node[output neuron] (O-1) at (3*\relusep + 2*\WSsep + 0.3cm,-2) {$x^3_0$};

    % Connect every node in the hidden layer with the output layer 
    \draw[nnedge] (I-1) --node[above,pos=0.7] {$1$} (H-1);
    \draw[nnedge] (I-2) --node[above,pos=0.7] {$1$} (H-1);
    % \draw[nnedge] (I-3) --node[below,pos=0.9] {$-1$} (H-1);
    \draw[nnedge] (I-1) --node[above,pos=0.8] {$1$} (H-2);
    \draw[nnedge] (I-2) --node[above,pos=0.7] {$-1$} (H-2);
    % \draw[nnedge] (I-3) --node[above,pos=0.8] {$1$} (H-2);
    % \draw[nnedge] (I-1) --node[above,pos=0.85] {$-1$} (H-3);
    % \draw[nnedge] (I-2) --node[above,pos=0.7] {$1$} (H-3);
    \draw[nnedge] (I-3) --node[above,pos=0.7] {$1$} (H-3);
    
    \draw[nnedge] (H-1) --node[above] {$\relu$} (H-4);
    \draw[nnedge] (H-2) --node[below] {$\relu$} (H-5);
    \draw[nnedge] (H-3) --node[below] {$\relu$} (H-6);
    
    \draw[nnedge] (H-4) --node[above,pos=0.7] {$1$} (H-7);
    \draw[nnedge] (H-5) --node[above,pos=0.7] {$1$} (H-7);
    % \draw[nnedge] (H-6) --node[below,pos=0.9] {$-1$} (H-7);
    \draw[nnedge] (H-4) --node[above,pos=0.75] {$-1$} (H-8);
    \draw[nnedge] (H-5) --node[above,pos=0.7] {$1$} (H-8);
    \draw[nnedge] (H-6) --node[above,pos=0.8] {$1$} (H-8);
    \draw[nnedge] (H-4) --node[above,pos=0.85] {$\ \ -1$} (H-9);
    \draw[nnedge] (H-5) --node[above,pos=0.7] {$1$} (H-9);
    \draw[nnedge] (H-6) --node[above,pos=0.7] {$-1$} (H-9);

    \draw[nnedge] (H-7) --node[below] {$\relu$} (H-10);
    \draw[nnedge] (H-8) --node[below] {$\relu$} (H-11);
    \draw[nnedge] (H-9) --node[below] {$\relu$} (H-12);
    
    \draw[nnedge] (H-10) --node[above,pos=0.5] {$1$} (O-1);
    \draw[nnedge] (H-11) --node[below,pos=0.5] {$1$} (O-1);
    \draw[nnedge] (H-12) --node[below,pos=0.5] {$1$} (O-1);
    
    % biases
    % \node[above right=-0.1cm and -0.2cm of H-9] (b1) {$-2$};
    % \node[below left=0.1cm and -0.25cm of H-5] (b2) {$-1$};

    % ranges
    \node[above=0.05cm of I-1] (r1) {$[-1,1]$};
    \node[above=0.05cm of I-2] (r2) {$[-1,1]$};
    \node[below=0.05cm of I-3] (r3) {$[-1,1]$};
    \node[above=0.05cm of H-1] (r4) {$[-2,2]$};
    \node[above=0.05cm of H-2] (r5) {$[-2,2]$};
    \node[below=0.05cm of H-3] (r6) {$[-1,1]$};
    \node[above=0.05cm of H-4] (r7) {$[0,2]$};
    \node[above=0.05cm of H-5] (r8) {$[0,2]$};
    \node[below=0.05cm of H-6] (r9) {$[0,1]$};
    \node[above=0.05cm of H-7] (r10) {$[0,2]$};
    \node[above=0.05cm of H-8] (r11) {$[-2,3]$};
    \node[below=0.05cm of H-9] (r12) {$[-3,2]$};
    \node[above=0.05cm of H-10] (r13) {$[0,2]$};
    \node[above=0.05cm of H-11] (r14) {$[0,3]$};
    \node[below=0.05cm of H-12] (r15) {$[0,2]$};
    \node[below=0.05cm of O-1] (r16) {$[0,6]$};
    
  \end{tikzpicture}
  }
  \caption{A neural network.}
  \label{fig:ExampleDnn}
\end{figure}
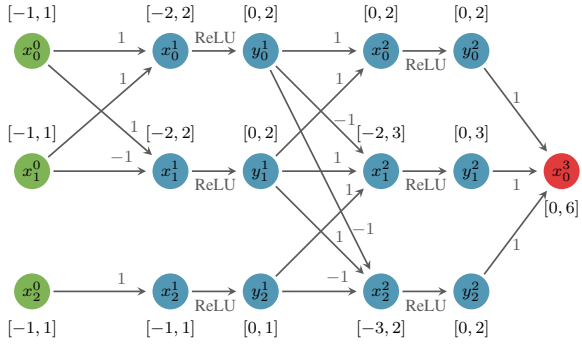

Before formally defining symbolic bound propagation, we start with an
intuitive example using the network in Fig.~\ref{fig:ExampleDnn}.  Let
$\vecx^i$ denote the \emph{pre-activation} values of the neurons in layer
$i$, and let $\boldsymbol{y}^i = \fsig{\vecx^i}$ denote their
\emph{post-activation} values; similarly, let $x^i_j$ and
$y^i_j = \fsig{x^i_j}$ denote the pre- and post-activation values of
neuron $j$ in layer $i$; and let $l^i_j, u^i_j$ denote the concrete
(scalar) lower- and upper-bound for $x^i_j$, i.e.
$l^i_j\leq x^i_j \leq u^i_j$ when the DNN is evaluated on any input
from $\inputDomain$. Assume that $\inputDomain$ is the following box
domain:
\[
  \inputDomain = \{-1 \leq x^0_i \leq 1 \ | \ i\in\{0,1,2\}\}
\]
and that we wish to compute bounds for the single output neuron, $x^3_0$.

We begin by propagating the bounds through the first affine
layer. According to the network's weights and biases, we get:
\[
    x^1_0 = x^0_0 + x^0_1,\qquad
    x^1_1 = x^0_0 - x^0_1,\qquad
    x^1_2 = x^0_2
\]
these equations allow us to compute concrete lower and upper bounds
for each of these neurons, by substituting the input neurons ($x^0_0,x^0_1,x^0_2$) with
their corresponding concrete bounds (according to the sign of their
coefficients). Using this process, we obtain:
\[
    x^1_0 \in [-2, 2],\qquad
    x^1_1 \in [-2, 2],\qquad
    x^1_2 \in [-1, 1]
\]
this propagation, often referred to as \emph{interval
  arithmetic}~\cite{Eh17}, is precise for individual neurons: indeed,
$x^1_0,x^1_1$ and $x^1_2$ can each take on any value in their
respective computed ranges. However, much important information is
lost when using just interval arithmetic: for example, it is
impossible for $x^1_0$ and $x^1_1$ to \emph{simultaneously} be
assigned $2$. As we will later see, symbolic-bound propagation
addresses this issue by capturing some of the dependencies between
neurons, and using these dependencies in producing tighter bounds.

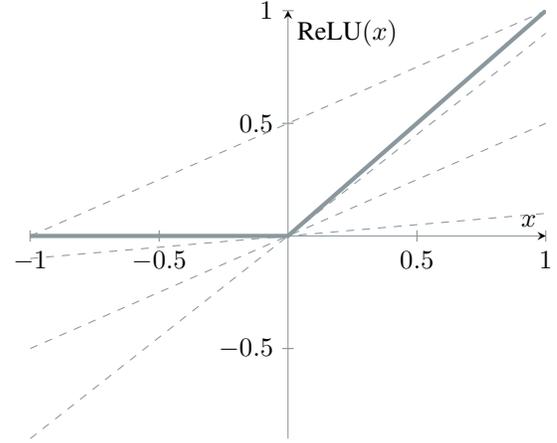
\begin{figure}[htp]
    \centering
    \begin{tikzpicture}[scale=1]
        \begin{axis}[domain=-1:1, axis lines=middle, xlabel=$x$, ylabel=$\frelu{x}$,]
            \addplot[style={ultra thick},] (x,{(x>=0) * x + (x<0) * 0)});
            \addplot[red, dashed] {0.5*(x+1)};
            \addplot[red, dashed] {0.1*x};
            \addplot[red, dashed] {0.5*x};
            \addplot[red, dashed] {0.9*x};
        \end{axis}
    \end{tikzpicture}
    \caption{A triangle relaxation of a ReLU function for $x\in[-1,1]$. The solid lines
      correspond to the exact ReLU function, and the dotted lines
      represent the relaxed lower and upper bounds, for different
      values of $\alpha$.}
    \label{fig:triangleRelaxation}
\end{figure}

For now, we continue propagating our computed bounds to neurons 
 $y^1_0$, $y^1_1$  and $y^1_2$. The output range of a \relu{} is the
 non-negative part of its input range, which yields:
\[
    y^1_0 \in [0, 2], \qquad
    y^1_1 \in [0, 2], \qquad
    y^1_2 \in [0, 1]
\]
and the next, affine layer is again handled using interval
arithmetic. Using the expressions
\[
    x^2_0 = y^1_0 + y^1_1,\qquad
    x^2_1 = -y^1_0 + y^1_1 + y^0_2,\qquad
    x^2_2 = -y^1_0 + y^1_1 - y^0_2
\]
% \begin{align*}
%     x^2_0 &= y^1_0 + y^1_1 = \fsig{x^1_0} + \fsig{x^1_1}\\
%     x^2_1 &= -y^1_0 + y^1_1 + y^0_2 = -\fsig{x^1_0} + \fsig{x^1_1} + \fsig{x^1_2}\\
%     x^2_2 &= -y^1_0 + y^1_1 - y^0_2 = -\fsig{x^1_0} + \fsig{x^1_1} - \fsig{x^1_2}
% \end{align*}
and substituting each $y^1_i$ with the appropriate bound, we obtain:
\[
  x^2_0\in [0,4],\qquad
  x^2_1\in [-2,4],\qquad
  x^2_0\in [-4,2]
\]
Unfortunately, as we soon show, the bounds computed for $x^2_0,x^2_1,x^2_2$
are not tight.
%This is a result of over-approximating the non-linear $\relu$
%activation layer that we passed through. Consequently,
A better approach is to compute \emph{symbolic bounds}, as opposed to
concrete ones, in a way that lets us carry additional information
about the dependencies between neurons. In symbolic-bound propagation,
we seek to express the upper and lower bounds of each neuron as a
linear combination of neurons from earlier layers, using a process
known as \emph{back-substitution}. The main difficulty is to propagate
these bounds across \relu{} layers, which are not convex; and this is
performed by using a \emph{triangle relaxation} of
the \relu{} function, illustrated in
Fig.~\ref{fig:triangleRelaxation}. Assume $x\in\left[l, u\right]$;
then, using this relaxation, we can deduce the following bounds:
\[
\begin{cases} 
    0 \leq \fsig{x} \leq 0 & \text{if\ } u\leq 0 \\
    x \leq \fsig{x} \leq x & \text{if\ } l\geq 0 \\
    \alpha x \leq \fsig{x} \leq \frac{u}{u-l}\left(x-l\right) &
    \text{otherwise, for any\ } 0\leq \alpha \leq 1 \\
\end{cases}
\]
Different symbolic bound propagation methods use different heuristics
for choosing $\alpha$~\cite{GaGePuVe19, XuZhWaWaJaLiHs20}; but this is
beyond our scope here, and our proposed technique is compatible with
any such heuristic. For our running example, we arbitrarily choose the
values of $\alpha$; and for our implementation, we use an existing
heuristic~\cite{XuZhWaWaJaLiHs20}.

% \begin{align*}
%     x^2_0 &= y^1_0 + y^1_1 = \fsig{x^1_0} + \fsig{x^1_1}\\
%     x^2_1 &= -y^1_0 + y^1_1 + y^0_2 = -\fsig{x^1_0} + \fsig{x^1_1} + \fsig{x^1_2}\\
%     x^2_2 &= -y^1_0 + y^1_1 - y^0_2 = -\fsig{x^1_0} + \fsig{x^1_1} - \fsig{x^1_2}
% \end{align*}

Using this relaxation, we show how to compute symbolic bounds that
yield tighter bounds for the $x^2_i$ neurons. First observe neuron
$x^2_0$, given as
$x^2_0 = y^1_0 + y^1_1 = \fsig{x^1_0} + \fsig{x^1_1}$.
To obtain its lower bound we first substitute both
$y^1_0 = \fsig{x^1_0}$ and $y^1_1 = \fsig{x^1_1}$ with their
corresponding triangle relaxation lower bounds, with the choice of
$\alpha = 0$ for both (we note that it is possible to choose different
$\alpha$ values for different variables). For the upper bound, we use
the linear upper bound from the triangle relaxation.  By using the
bounds we already know for nodes in previous layers, we get that:
\begin{align*}
  x^2_0 &\geq 0 \cdot x^1_0 + 0 \cdot x^1_1 = 0 \\
    x^2_0 &\leq \frac{1}{2}\left(x^1_0 + 2\right) + \frac{1}{2}\left(x^1_1 + 2\right) = \frac{1}{2}\left(x^1_0 + x^1_1\right) + 2\\
    &= \frac{1}{2}\left(\left(x^0_0 + x^0_1\right) + \left(x^0_0 -
      x^0_1\right)\right) + 2 = x^0_0 + 2 \leq 3\\
\end{align*}
which indeed produces a tighter upper bound than the one obtained for
$x^2_0$ using interval propagation.  Similarly, we get that for
$x_1^2$:
\begin{align*}
    x^2_1 &\geq -\frac{1}{2}\left(x^1_0 + 2\right) + 0 \cdot x^1_1 + 0\cdot x^1_2\\
    &= -\frac{1}{2}\left(x^0_0 + x^0_1\right) - 1 = -2\\
    x^2_1 &\leq - 0 \cdot x^1_0 + \frac{1}{2}\left(x^1_1 + 2\right) + \frac{1}{2}\left(x^1_2 + 1\right)\\
    &= \frac{1}{2}\left(x^1_1 + x^1_2\right) + 1.5 =
      \frac{1}{2}\left(x^0_0 - x^0_1 + x^0_2\right) + 1.5  \leq 3
\end{align*}
and for $x_2^2$:
\begin{align*}
    x^2_2 &\geq -\frac{1}{2}\left(x^1_0 + 2\right) + 0 \cdot\left(x^1_1\right) - \frac{1}{2}\left(x^1_2 + 1\right)\\
    &= -\frac{1}{2}\left(x^1_0 + x^1_2\right) - 1.5 = -\frac{1}{2}\left(x^0_0 + x^0_1 + x^0_2\right) - 1.5 \geq -3\\
    x^2_2 &\leq - 0 \cdot x^1_0 + \frac{1}{2}\left(x^1_1 + 2\right) - 0 \cdot x^1_2\\
    &= \frac{1}{2}x^1_1 + 1 = \frac{1}{2}\left(x^0_0 - x^0_1\right) + 1 \leq 2
\end{align*}
We have thus obtained the following bounds:
\[
  x^2_0 \in [0, 3], \qquad
    x^2_1 \in [-2, 3], \qquad
    x^2_2 \in [-3, 2]
\]
We note that while these bounds are tighter than the ones produced by
interval propagation, and are in fact optimal for $x^2_1,x^2_2$, this
is not the case for $x^2_0$ (the optimal bounds are displayed in
square brackets in Fig.~\ref{fig:ExampleDnn}). The reason for this
sub-optimality is discussed in Section~\ref{sec:errorsinBS}.

We continue to propagate our bounds through the next layer, obtaining:
\[
    y^2_0 \in [0, 3],\qquad
    y^2_1 \in [0, 3],\qquad
    y^2_2 \in [0, 2]
  \]
  % y^2_0 &= \fsig{x^2_0} &\in [0, 3]\\
  % y^2_1 &= \fsig{x^2_1} &\in [0, 3]\\
  %   y^2_2 &= \fsig{x^2_2} &\in [0, 2]
and finally reach:
\begin{align*}
    x^3_0 &= y^2_0 + y^2_1 + y^2_2
    = \fsig{x^2_0} + \fsig{x^2_1} + \fsig{x^2_2}\\
    &\leq x^2_0 + \frac{3}{5}\left(x^2_1 + 2\right) + \frac{2}{5}\left(x^2_2 + 3\right)\\
    % &= \begin{aligned}[t]
    %         \left(y^1_0 + y^1_1\right) &+ \frac{3}{5}\left(-y^1_0 + y^1_1 + y^1_2\right)\\
    %                                    &+ \frac{2}{5}\left(-y^1_0 + y^1_1 - y^1_2\right) + \frac{12}{5}
    % \end{aligned}\\
    &= 2y^1_1 + \frac{1}{5}y^1_2 + \frac{12}{5}
    = 2\fsig{x^1_1} + \frac{1}{5}\fsig{x^1_2} + \frac{12}{5}\\
    &\leq 2 \cdot \frac{1}{2}\left(x^1_1 + 2\right) + \frac{1}{5} \cdot \frac{1}{2}\left(x^1_2 + 1\right) + \frac{12}{5}\\
    &= x^0_0 - x^0_1 + \frac{1}{10}x^0_2 + 4.5 \leq 6.6
\end{align*}

% backsub
More generally, the back-substitution process for upper-bounding a
neuron $x^k_i$ (assuming we already have valid bounds for all neurons
in earlier layers) is iteratively defined as:
\begin{align*}
    \fmax{x^k_i} & = \fmax{W^{k-1}_{i}\fsig{\vecx^{k-1}}} \\
    & \leq \fmax{W^{k-1}_{i}R^{k-2}_U\vecx^{k-1}} \\
    & = \fmax{W^{k-1}_{i}R^{k-2}_U W^{k-2}\fsig{\vecx^{k-2}}} \\
    & \leq \fmax{W^{k-1}_{i}R^{k-2}_UW^{k-2}R^{k-3}_U\vecx^{k-2}} \\
    & = \ldots \leq \fmax{W^{k-1}_{i} \prod_{j=k-2}^{0}\left(R^{j}_UW^{j}\right) \vecx^{0}}
\end{align*}
(Biases and constants are handled similarly, and are omitted for
clarity.) At each step, we can replace the variables of $\vecx^{i}$ by their
respective concrete bounds $[l^i_j, u^i_j]$, in an interval-arithmetic
fashion, to obtain a valid concrete upper bound for the value of
$\fmax{x^k_i}$. We refer to this operation as
\emph{concretization}. We call the matrices $R^i_L,R^i_U$ the
respective lower- and upper-bound \emph{relaxation}
matrices~\cite{XuZhWaWaJaLiHs20}. These matrices apply the appropriate
triangle relaxation to each $\relu$, allowing us to replace it with a
linear bound, and are defined using the current symbolic bounds for
each $\relu$ as well as the weight matrix of the layer the precedes
it. The two matrices are defined such that
$\forall \vecx \in \inputDomain$:
\[\boldsymbol{\omega_i} R^i_L \vecx + c_L\leq \boldsymbol{\omega_i}\fsig{\vecx} \leq \boldsymbol{\omega_i} R^i_U \vecx + c_U\]
where $c_L$ and $c_U$ are scalar constants; and $\boldsymbol{\omega_i}$ is a row
vector containing the coefficients of each $\fsig{x_j}$, resulting in
linear bounds for the sum of $\relu$s.  A precise definition of these
matrices appears in Sec.~\ref{sec:appendix:relaxationMatrices} of the
Appendix; and a similar procedure can be applied for lower-bounding
$x^k_i$.

At first glance, the iterative back-substitution process may seem
counter productive; indeed, in each iteration where we move to an
earlier layer of the network, we use a less-than-equals transition,
which seems to indicate that the upper bound that we will eventually
reach is more loose than the present bound. This, however, is not so;
and the reason is the \emph{concretization} process. When we
concretize the bounds in some later iteration, it is possible that the
known bounds for the variables in that layer of the network will lead
to a tighter upper bound than the one that can be derived
presently. More generally, this process can be regarded as a trade-off between computing
looser expressions for the bound, but being able to concretize them
over more exact domains --- which could result in tighter
bounds~\cite{GaGePuVe19}.
%\tom{this is also true for the lower bound but with a greater-than-equals transition}
 
\section{Errors in Back-Substitution}
\label{sec:errorsinBS}
%\subsection{Back-substitution with errors}

As previously mentioned, although symbolic-bound computation using
back-substitution can derive tighter bounds than na\"ive interval
propagation, there are cases in which the computed bounds are
sub-optimal: for example, while the bounds computed for $x^2_1$ and
$x^2_2$ were tight (i.e., there exists an input in $\inputDomain$ for
which they are met), the bounds for $x^2_0$ and $x^3_0$ were not. In
this section, we analyze the reasons behind such sub-optimal
bounds. %, in order to later address them.
% A natural question that arises is why are
% the concrete-bounds found for $x^2_1,x^2_2$ are optimal (in the sense
% that there are inputs within $\inputDomain$ for which each neuron
% obtains the value of its concrete-bounds) while those found for
% $x^2_0, x^3_0$ are not (there is no input in $\inputDomain$ for which
% $x^2_0 > 2$ or $x^3_0 > 6$)? despite both being the result of the same
% over-approximation process that contains the same type of
% over-approximation errors?  In this section we analyze the way errors
% are created and accumulated during the back-substitution process and
% attempt to answer this question.
% This is a result of the over-approximation we performed during back-substitution, specifically when we replaced $\relu$ functions with their corresponding lower/upper bounds. We define:
We begin with the following definitions:
\begin{definition}[Optimal bias for bound]
%   Let $f:\mathbb{R}^{n}\rightarrow\mathbb{R}$, and let
%   $\omega_U$ be \guy{TODO: what is omega u?}. We say that $b_U$ is the
%   optimal bias for $f$'s upper bound if it is the smallest scalar for
%   which $L_U\geq f(x)$, where
%   $L_U\equiv (\boldsymbol{x})\equiv \omega_U \boldsymbol{x} + b_L$.
%   The definition for the optimal bias for $f$'s lower bound, $b_L$, is
%   symmetrical. 
  let $f:\mathbb{R}^{n}\rightarrow\mathbb{R}$ be a function and let
  $U_{f}(\boldsymbol{x}) \equiv \boldsymbol{\omega} \boldsymbol{x} +
  b$ ($\boldsymbol{\omega} \in \mathbb{R}^n,\, b\in \mathbb{R}$) be a
  valid linear upper bound for $f$ over the domain $\mathcal{D}$,
  i.e.,
  $\forall \boldsymbol{x}\in\mathcal{D}:U_{f}(\boldsymbol{x})\geq
  f(\boldsymbol{x})$.  We say that $b$ is the \emph{optimal bias} for
  $U_f(\boldsymbol{x})$ if $\forall b^*:b^*<b$, it holds that
  $U^*_{f}(\boldsymbol{x}) \equiv \boldsymbol{\omega} \boldsymbol{x} +
  b^*$ is no longer a valid upper bound for $f$. The definition
  for the optimal bias for $f$'s lower bound is symmetrical.
% let $f:\mathbb{R}^{n}\rightarrow\mathbb{R}$ and
%   $L_{f}\left(\boldsymbol{x}\right)\equiv \omega_L \boldsymbol{x} +
%   b_L$, be valid lower bound and
%   $U_{f}\left(\boldsymbol{x}\right)\equiv \omega_U \boldsymbol{x} +
%   b_U$ be a valid upper bound for $f$ over the domain
%   $\mathcal{D}$. that is:
%   $\forall
%   \boldsymbol{x}\in\mathcal{D}:U_{f}\left(\boldsymbol{x}\right)\geq
%   f\left(\boldsymbol{x}\right) \geq
%   L_{f}\left(\boldsymbol{x}\right)$. then $b_L$ is called an
%   optimal-bias for the lower bound and $b_U$ is called an optimal-bias
%   for the upper bound if $\forall b^*_L:b^*_L>b$ and
%   $\forall b^*_U:b^*_U<b$, then
%   $L^{*}_{f}\left(\boldsymbol{x}\right) \equiv \omega_L \boldsymbol{x}
%   + b^{*}_L$ and
%   $U^{*}_{f}\left(\boldsymbol{x}\right) \equiv \omega_U \boldsymbol{x}
%   + b^{*}_U$ are no longer valid lower and upper bounds for $f$ an
%   example of an sub-optimal and optimal upper bounds are illustrated
%   in fig.~\ref{fig:errortypes}.
\end{definition}

\begin{figure}[ht]
    \centering
    \begin{tikzpicture}%[scale=0.59]
        \begin{axis}[domain=-1:1, axis lines=middle, xlabel=$x$, ylabel=$y$,]
            \addplot[color=blue,smooth, ultra thick,] {(x>=0) * x + (x<0) * 0)} node[below right, pos=0.7] {$\relu(x)$};
            \addplot[red, dashed] {0.5*(x+2)} node[above, pos=0.8] {$\frac{1}{2}x+1$};
            \draw[red, ->] (axis cs:0, 1) -- (axis cs:0, 0.49);
            \addplot[purple, dashed] {0.5*(x+1)} node[above, pos=0.8] {$\frac{1}{2}x+\frac{1}{2}$};;
        \end{axis}
    \end{tikzpicture}
    % \begin{tikzpicture}[scale=0.59]
    %     \begin{axis}[domain=-1:1, axis lines=middle, xlabel=$x$, ylabel=$y$,]
    %         \addplot[color=blue,smooth, ultra thick,] {(x>=0) * x + (x<0) * 0)} node[below right, pos=0.7] {$\relu(x)$};
    %         \addplot[red, dashed] {0.75*(x+1)} node[above, pos=0.8] {$\frac{3}{4}x+\frac{3}{4}$};
    %         \draw[red, ->] (axis cs:0.96, 1.47) -- (axis cs:0.96, 0.97);
    %         \addplot[purple, dashed] {0.5*(x+1)} node[above, pos=0.8] {$\frac{1}{2}x+\frac{1}{2}$};
    %     \end{axis}
    % \end{tikzpicture}
    % \caption{Simplified visualisation of two types of sub-optimal bounds caused by relaxation errors and their correction. The first is an sub-optimal bias and the second are sub-optimal slope}
    \caption{A simplified illustration of an optimal and sub-optimal bounds for a \relu{} function over $x\in[-1,1]$.}
    \label{fig:errortypes}
\end{figure}
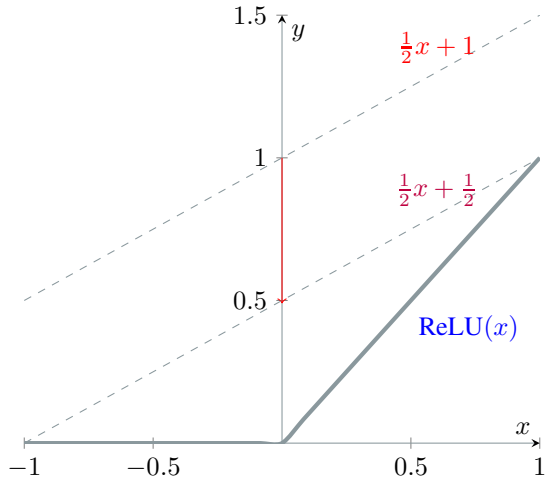

An example of optimal and sub-optimal upper bounds appears in
Fig.~\ref{fig:errortypes}.  In the graph depicted therein, we plot an
upper bound for the function $\relu(x)$. The
bias value of the first bound (in red) is 1; and as we can see, the
resulting bound is not tight. When we set the bias value to $1/2$, the
bound becomes tight, equaling the function at points $x=-1$ and $x=1$,
and so that is the optimal bias value for that bound.
% A similar
% situation occurs on the right hand side. \guy{I would change in figure
%   4 from 1/2(x+1) to 1/2x + 1/2. Also, the green font is almost
%   invisible, maybe change to purple or brown}

%\todo{Define the notion of an optimal slope for a bound: not as trivial, only sometimes applies, maybe define sub-optimal slope instead?
%maybe: $\forall \lambda \in \left[0,1\right) \lambda U_f$ is not a valid bound?}

\begin{definition}[Bound error]
Let $f:\mathbb{R}^{n}\rightarrow\mathbb{R}$, and let $g(\vecx)$ be an
upper bound for $f$ over domain $\mathcal{D}$, such that we have: $\forall
\vecx\in\mathcal{D}: g(\vecx)\geq f(\vecx)$. We define the error of $g$ with
respect to $f$ as the function:
$E(\vecx) = g(\vecx) - f(\vecx)$. The case for a lower bound is symmetrical.
%   The bound error of a function is the difference between the bound and the true value of the function:
% let $f:\mathbb{R}^{n}\rightarrow\mathbb{R}$ and $g\left(x\right)$ be a lower/upper such that 
% $\forall x\in\mathcal{D}: g\left(x\right) \leq f\left(x\right)$, for a lower bound
% or $\forall x\in\mathcal{D}: g\left(x\right) \geq f\left(x\right)$, for an upper bound
% The bound error is simply defined as:
% $E\left(x\right) = g\left(x\right) - f\left(x\right)$
\end{definition}

We observe that a linear bound $g$ for $f$ over the domain
$\inputDomain$ has \emph{optimal bias} iff
$\exists \vecx\in \inputDomain: E(\vecx) = 0$. We refer to any bound that has a
sub-optimal bias, i.e.  $\forall \vecx \in \inputDomain: E(\vecx) > 0$, as a
\emph{detached bound}. We show that these detachments occur naturally
as part of the back-substitution process, and are partially
responsible for the discovery of sub-optimal concrete bounds.

% We now give a short intuition on the way detached bounds form:

It is straightforward to see that the aforementioned triangle
relaxation for \relu{}s produces linear bounds that are bias-optimal
for each individual $\relu$. However, as it turns out, this may not be the case when multiple \relu{}s
are involved. In a typical DNN, a neuron's value is computed as a
weighted sum of the \relu{}s of values from its preceding layer.
Consequently, when we calculate an upper bound for the neuron using
back-substitution, we are in fact upper-bounding a sum of
\relu{}s by summing their individual upper bounds. This can result in a
\emph{detached bound}, where, despite the fact that each \relu{} was
approximated using a bound with an optimal bias, the resulting
combined bound does not have optimal bias.

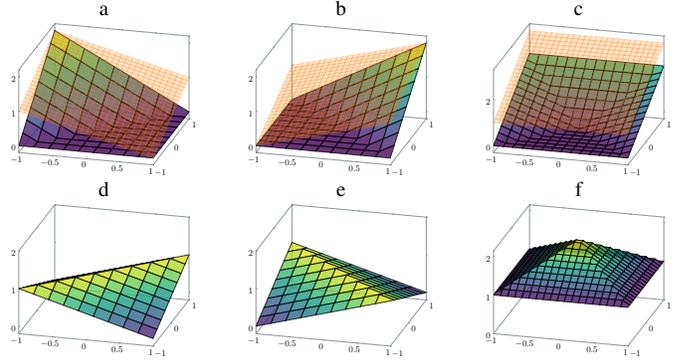
\begin{figure}[ht]
    \centering
    \begin{tikzpicture}[scale=0.33]
        \begin{axis}[shader=flat, tickwidth=0pt, view/h=15, title=a, title style={font=\Huge}]
            \addplot3[surf, samples=10, opacity=0.8, domain=-1:1, y domain=-1:1, colormap/viridis, draw=black]{max(0,-x+y)};
            \addplot3[surf, color=orange, opacity=0.2, domain=-1:1, y domain=-1:1,]{0.5*(-x+y+2)};
        \end{axis}
    \end{tikzpicture}
    \hfill
    \begin{tikzpicture}[scale=0.33]
        \begin{axis}[shader=flat, tickwidth=0pt, view/h=15, title=b, title style={font=\Huge}]
            \addplot3[surf, samples=10, opacity=0.8, domain=-1:1, y domain=-1:1, colormap/viridis, draw=black]{max(0,x+y)};
            \addplot3[surf, color=orange, opacity=0.2, domain=-1:1, y domain=-1:1,]{0.5*(x+y+2)};
        \end{axis}
    \end{tikzpicture}
    \hfill
    \begin{tikzpicture}[scale=0.33]
        \begin{axis}[shader=flat, tickwidth=0pt, view/h=15, title=c, title style={font=\Huge}]
            \addplot3[surf, samples=15, opacity=0.8, domain=-1:1, y domain=-1:1, colormap/viridis, draw=black]{max(0,-x+y) + max(0,x+y)};
            \addplot3[surf, color=orange, opacity=0.2, domain=-1:1, y domain=-1:1,]{y+2)};
        \end{axis}
    \end{tikzpicture}

    \begin{tikzpicture}[scale=0.33]
        \begin{axis}[shader=flat, tickwidth=0pt, view/h=15, zmax=2, title=d, title style={font=\Huge}]
            \addplot3[surf, samples=10, opacity=0.8, domain=-1:1, y domain=-1:1, colormap/viridis, draw=black]{0.5*(-x+y+2) - max(0,-x+y)};
        \end{axis}
    \end{tikzpicture}
    \hfill
    \begin{tikzpicture}[scale=0.33]
        \begin{axis}[shader=flat, tickwidth=0pt, view/h=15, zmax=2, title=e, title style={font=\Huge}]
            \addplot3[surf, samples=10, opacity=0.8, domain=-1:1, y domain=-1:1, colormap/viridis, draw=black]{0.5*(x+y+2) - max(0,x+y)};
        \end{axis}
    \end{tikzpicture}
    \hfill
    \begin{tikzpicture}[scale=0.33]
        \begin{axis}[shader=flat, tickwidth=0pt, view/h=15, zmin=0, title=f, title style={font=\Huge}]
            \addplot3[surf, samples=20, opacity=0.8, domain=-1:1, y domain=-1:1,, colormap/viridis, draw=black]{y+2 - max(0,-x+y) - max(0,x+y)};
        \end{axis}
    \end{tikzpicture}
    \caption{Illustration of the formation of detached bounds as a
      result of summed errors. Sub-figures $a$ and $b$ correspond to
      $y^1_0 = \frelu{x^0_0 + x^0_1}$, $y^1_1 = \frelu{x^0_0 - x^0_1}$
      and their relaxed upper bounds (in orange); and sub-figure $c$
      corresponds to $x^2_0 = y^1_0 + y^1_1$ and its symbolic upper bound,
      computed using back-substitution.}
    \label{fig:detachedBounds}
\end{figure}

An illustration of this phenomenon appears in
Fig.~\ref{fig:detachedBounds}. Sub-figures \emph{a} and \emph{b}
therein show the graph of $\relu$ functions, plotted along their
triangle-relaxation upper bound (in orange). Sub-figure \emph{c} then
shows the graph of the \emph{sum} of the two $\relu$ functions from
sub-figures \emph{a} and \emph{b}, along with the sum of their
individual upper bounds (again, in orange). As we can see, although
the upper bounds in \emph{a} and \emph{b} touch the functions they are
approximating in at least one point (and are hence bias-optimal), the
bound in $\emph{c}$ is detached, and is hence not bias-optimal. Each
figure in the lower row of Fig.~\ref{fig:detachedBounds} shows the
over-approximation error of the figure directly above it.
 
% We note that the bound produced by the "triangle" relaxation for a single $\relu$ for a bounded variable" $x\in\left[l,u\right]$ has an optimal bias and the error of each over-approximation $\frelu{x} \leq \frac{u}{u-l}\left(x-l\right)$ is exactly $0$ at both coordinates of $x=l$ and $x=u$.

More formally, the error of the upper bound for $\relu{}(x)$ with
current bounds $l<0<u$ is:
\[
  E(x) = \frac{u}{u-l}(x-l) - \fsig{x} \qquad x\in[l, u]
\]
% and for a lower-bound:
% \[ E\left(x\right) = \alpha x - \fsig{x} \]
and we note that $E(l) = E(u) = 0$.  In more complex cases, such as
the case of the multivariate function $x^2_0 = y^1_0 + y^1_1$ depicted
in Fig.~\ref{fig:detachedBounds}, the coordinates where the bound
error equals zero could be different for $y^1_0$ and $y^1_1$ ---
resulting in the bound obtained for $x^2_0$, their sum, becoming
detached from the true value of the function.  We now show it for the
case of $x^2_0$ in greater detail:
\[
  x^2_0 = \fsig{x^1_0} + \fsig{x^1_1} = \fsig{x^0_0 + x^0_1} +
  \fsig{x^0_0 - x^0_1}
\]
An upper bound is computed using the relaxations:
\begin{align*}
\fsig{x^0_0 + x^0_1} \leq \frac{1}{2}\left(x^0_0 + x^0_1 + 2\right)\\
\fsig{x^0_0 - x^0_1} \leq \frac{1}{2}\left(x^0_0 + x^0_1 + 2\right)
\end{align*}
where each relaxation has its own relaxation error:
\begin{align*}
E^1_0(x^0_0, x^0_1) = \frac{1}{2}(x^0_0 + x^0_1 + 2) - \fsig{x^0_0 + x^0_1}\\
E^1_1(x^0_0, x^0_1) = \frac{1}{2}(x^0_0 + x^0_1 + 2) - \fsig{x^0_0 - x^0_1}
\end{align*}
The relaxed linear bound obtained is:
\[
  x^2_0 \leq \frac{1}{2}(x^0_0 + x^0_1 + 2) +
  \frac{1}{2}(x^0_0 + x^0_1 + 2) = x^0_0 + 2
\] 
And its error is the sum of the errors of its summands:
\begin{align*}
  E_\text{total}(x^0_0, x^0_1) &\equiv E^1_0 + E^1_1 \\
  &=x^0_0 + 2 -  \fsig{x^0_0 + x^0_1} - \fsig{x^0_0 - x^0_1}
\end{align*}
We note that:
\begin{align*}
\fmin{E^1_0} = E^1_0(-1, -1) = E^1_0(1, 1) = 0\\
\fmin{E^1_1} = E^1_1(-1, 1) = E^1_1(1, -1) = 0
\end{align*}
However:
\[\fmin{E_\text{total}} = E_\text{total}\left(-1, x^0_1\right) = 1\]
The reason for this is that at the coordinates $\langle -1, -1\rangle$
and $\langle 1, 1\rangle$ where
$E^1_0\left(-1,-1\right) = E^1_0\left(1,1\right) = 0$, we have that
$E^1_1\left(-1,-1\right) = E^1_1\left(1,1\right) = 1$; and vice-versa,
for the coordinates $\langle -1, 1\rangle$ and $\langle 1, -1\rangle$,
where $E^1_1\left(-1,1\right) = E^1_1\left(1,-1\right) = 0$ and
$E^1_0\left(-1, 1\right) = E^1_0\left(1, -1\right) = 1$.  The
optimal linear bound for
\[
  x^2_0 = \fsig{x^0_0 + x^0_1} + \fsig{x^0_0 - x^0_1}
\]
is in fact $x^2_0 \leq x^0_0 + 1$, which is the bias-optimal version of the
existing linear bound of $x^2_0 \leq x^0_0 + 2$.

\section{DeepMIP: Minimizing Back-Substitution Errors}
\label{sec:DeepMIP}

During a back-propagation execution, the over-approximations of
individual \relu{}s are repeatedly summed up, which leads to bounds
that become increasingly more detached with each iteration --- and
this results in very loose concrete bounds that hamper
verification. We now describe our method, which we term \emph{DeepMIP},
for ``tightening'' detached bounds, with the goal of eventually
obtaining tighter concrete bounds.
The idea is to alter the back-propagation mechanism, so that in each
iteration it \emph{minimizes} the sum of errors that result from the
relaxation of the current activation layer --- effectively pushing loose
upper bounds down towards the function, by decreasing their bias
values (a symmetrical mechanism can be applied for lower bounds).
More specifically, we propose to rewrite the general back-substitution
rule for a single iteration as follows:
\begin{align*}
    \fmax{x^k_i} & = \fmax{W^{k-1}_i\fsig{\vecx^{k-1}}}\\
    &= \max \big( W^{k-1}_i R^{k-2}_U \vecx^{k-1}\\
    &\qquad\qquad- \left(W^{k-1}_i R^{k-2}_U \vecx^{k-1} - W^{k-1}_i  \fsig{\vecx^{k-1}} \right)\big)\\
    &= \fmax{W^{k-1}_i R^{k-2}_U \vecx^{k-1} - E^{k-1}}\\
    &\leq \fmax{W^{k-1}_i R^{k-2}_U\vecx^{k-1}} - \fmin{E^{k-1}}
\end{align*}
Observe that while $\fmin{E^{k-1}}$ is non-convex, it contains no
nested $\relu$s, and can often be efficiently solved by MIP
solvers~\cite{TjXiTe17}. Thus, as DeepMIP performs the iterative
back-substitution process, it can invoke a MIP solver to minimize the
error in each iteration, and use it to improve the deduced bounds. The
pseudo-code for the algorithm appears in Sec.~\ref{sec:appendix:pseudocode} of
the Appendix. Observe that MiniMIP can be regarded as a generalization
of modern back-substitution methods~\cite{XuZhWaWaJaLiHs20,
  GaGePuVe19}, in the sense that they only use the non-negativity of
the error to produce a trivial bound:
\[
  \fmin{E^{k-1}} = \fmin{W^{k-1}_{i} R^{k-2}_U \vecx^{k-1} -
    W^{k-1}_{i}\fsig{\vecx^{k-1}}} \geq 0
\]
which is correct, since the error of an upper bound is non-negative by
definition (in the lower bound case, the error is non-positive, and so $0$
can be used as a trivial upper bound).

To continue our computation we denote the error caused by the over-approximation of the activation of layer $t$ during back-substitution as:
\begin{equation}\label{errordefinition}
E^t \equiv W^{k-1}_{i} \prod_{j=k-2}^{t-1}(R^{j}_U W^{j})(R^{t-1}_U \vecx^{t} - \fsig{\vecx^{t}})
\end{equation}

In the definition above, $i$ is the index of the neuron being bounded by the back-substitution. 
We get:
 \begin{align*}
   \fmax{x^k_i} &\leq \fmax{W^{k-1}_i R^{k-2}_U \vecx^{k-1}} - \fmin{E^{k-1}}\\
    &= \fmax{W^{k-1}_i R^{k-2}_U W^{k-2}\fsig{\vecx^{k-2}}} - \fmin{E^{k-1}}\\
    &= \fmax{W^{k-1}_i R^{k-2}_U W^{k-2} R^{k-2}_U \vecx^{k-2} - E^{k-2}}\\
  &\textcolor{white}{=} - \fmin{E^{k-1}}\\
    &\leq \fmax{W^{k-1}_i R^{k-2}_U W^{k-2} R^{k-2}_U
      \vecx^{k-2}} \\
  &\textcolor{white}{=} - \fmin{E^{k-2}} - \fmin{E^{k-1}}\\
               &= \ldots \\
  &\leq \fmax{W^{k-1}_{i} \prod_{j=k-2}^{0}(R^{j}_UW^{j})
    \vecx^{0}} - \sum^0_{j=k-1}\fmin{E^j}
\end{align*}
Finally, the maximization problem is transformed into a linear sum over a
box domain, which is easy to solve.
 Since each $E^j$ is shallow (contains no nested $\relu$s), it can be
minimized efficiently using MIP solvers, and each non-trivial minimum that
is found will improve the tightness of the final upper bound. %(and as mentioned before, most modern
%neural-network verification tools are willing to take this tradeoff).
However, we note that the number of MIP problems generated by this
process increases linearly with the \emph{depth} of the neuron within
the network --- i.e., for a neuron in layer $k$, there are $k$
minimization problems to solve.  For deeper networks, especially ones
with large domains or ones where many layers only have very loose
bounds, minimizing the error terms could become computationally
expensive.

\mysubsection{Optimization: Direct MIP encoding.}
As part of its operation, DeepMIP
dispatches MIP problems, each corresponding to the
over-approximation error of a particular layer.  Specifically when it
over-approximates the first layer:
\begin{align*}
    &\fmax{W^{k-1}_{i} \prod_{j=k-2}^{1}(R^{j}_UW^{j})\fsig{W^0 \vecx^{0}}} - \sum_{j=k-2}^{1}\fmin{E^j}\\
    &\leq \fmax{W^{k-1}_{i} \prod_{j=k-2}^{0}(R^{j}_UW^{j})\vecx^{0}} - \fmin{E^0}\\
    &\qquad- \sum_{j=k-2}^{1}\fmin{E^j}\\
\end{align*}
it will directly solve the linear optimization problem:
\[ \fmax{W^{k-1}_{i} \prod_{j=k-2}^{0}(R^{j}_UW^{j})\vecx^{0}} \]
and use a MIP solver to solve:
\[
  \fmin{E^0}=\min \bigg(W^{k-1}_{i} \prod_{j=k-2}^{1}(R^{j}_U W^{j})(R^{0}_U \vecx^{t} - \fsig{\vecx^{0}}\bigg) 
\]
We observe that in this particular case, since we reached the input layer, the initial term can instead be directly solved as a separate MIP query: 
\[
\fmax{W^{k-1}_{i} \prod_{j=k-2}^{1}(R^{j}_UW^{j})\fsig{W^0 \vecx^{0}}}
\]
which may result in tighter bounds, since it prevents any additional
imprecision.
%which may be introduced by using approximations with sub-optimal slopes.
%(illustrated in sub-figure \textbf{b} in fig.~\ref{fig:errortypes})
We note that this optimization to DeepMIP generalizes the common
practice of directly finding the concrete bounds of the neurons in the
first layer using MIP solvers, and only applying back-substitution
from the second layer onward~\cite{MuMaSiPuVe22, XuZhWaWaJaLiHs20}.

We illustrate this approach by repeating the back-substitution process
for $x^3_0$ from our running example:
\begin{align*}
    \fmax{x^3_0} &= \fmax{y^2_0 + y^2_1 + y^2_2}\\
    &= \fmax{\fsig{x^2_0} + \fsig{x^2_1} + \fsig{x^2_2}}\\
    &= \max\bigg(\fsig{y^1_0 + y^1_1} + \fsig{-y^1_0 + y^1_1 + y^1_2}\\
        &\qquad\qquad+ \fsig{-y^1_0 + y^1_1 - y^1_2}\bigg)\\
    % &= x^2_0 + \frac{3}{5}\left(x^2_1 + 2\right) + \frac{2}{5}\left(x^2_2 + 3\right)\\
    % &\qquad- \left(x^2_0 + \frac{3}{5}\left(x^2_1 + 2\right) +
    % \frac{2}{5}\left(x^2_2 + 3\right) - \fsig{x^2_0} - \fsig{x^2_1}
    % - \fsig{x^2_2}\right)\\
                 &= \max(A - E^2_U) \leq \max(A) - \min(E^2_U)
\end{align*}
where
\begin{multline*}
  A = (y^1_0 + y^1_1) + \frac{3}{5}(-y^1_0 + y^1_1 + y^1_2)
  + \frac{2}{5}(-y^1_0 + y^1_1 - y^1_2) + \frac{12}{5}\\
  = 2y^1_1 + \frac{1}{5}y^1_2 + \frac{12}{5}
\end{multline*}
and $E^2_U$ is defined as per Eq.~\ref{errordefinition}:
\begin{align*}
  E^2_U &= (y^1_0 + y^1_1) + \frac{3}{5}(-y^1_0 + y^1_1 + y^1_2) \\
    &\quad+ \frac{2}{5}(-y^1_0 + y^1_1 - y^1_2) + \frac{12}{5} - \fsig{y^1_0 + y^1_1}\\
    &\quad-\fsig{-y^1_0 + y^1_1 + y^1_2} - \fsig{-y^1_0 + y^1_1 - y^1_2}\\
    &= 2y^1_1 + \frac{1}{5}y^1_2 + \frac{12}{5} - \fsig{y^1_0 + y^1_1}\\
    &\quad-\fsig{-y^1_0 + y^1_1 + y^1_2} - \fsig{-y^1_0 + y^1_1 - y^1_2}
\end{align*}

Simplifying these expressions, we get that
\begin{align*}
  \fmax{x^3_0}
  &\leq \max(A) - \min(E^2_U) \\
  &= \max(2y^1_1 + \frac{1}{5}y^1_2 + \frac{12}{5}) -\min(E^2_U)
\end{align*}
Using a MIP solver to find the minimum of $E^2_U$ over the variables
of $\boldsymbol{y}^1$ reveals that  $\min(E^2_U) =
\frac{2}{5}$. We substitute this, and get:
\begin{align*}
  \fmax{x^3_0}
  &\leq \max(2y^1_1 + \frac{1}{5}y^1_2 + \frac{12}{5}) -\frac{2}{5}
\end{align*}
Finally, since we have reached the first layer, we write:
\begin{align*}
  \fmax{x^3_0}
  &\leq \max(2y^1_1 + \frac{1}{5}y^1_2 + \frac{12}{5}) -\frac{2}{5} \\
  &= \fmax{2\fsig{x^1_1} + \frac{1}{5}\fsig{x^1_2} + \frac{12}{5}} - \frac{2}{5}\\
  &= \fmax{2\fsig{x^0_0 - x^0_1} + \frac{1}{5}\fsig{x^0_2} +
    \frac{12}{5}} - \frac{2}{5}
\end{align*}
and then, using our proposed enhancement, we directly solve this maximization over the input layer instead of back-substituting it any further. The MIP solver replies that:
\[
  \fmax{2\fsig{x^0_0 - x^0_1} + \frac{1}{5}\fsig{x^0_2} +
  \frac{12}{5}} = 6\frac{2}{5}
\]
and we then substitute this value to obtain:

\[
  \fmax{x^3_0} \leq 6\frac{2}{5} - \frac{2}{5} = 6
\]

%   \begin{align*}
%     &= \begin{aligned}[t]
%         \max\Bigg(&\left(y^1_0 + y^1_1\right) + \frac{3}{5}\left(-y^1_0 + y^1_1 + y^1_2\right)\\
%             &+ \frac{2}{5}\left(-y^1_0 + y^1_1 - y^1_2\right) + \frac{12}{5}\\
%             &-\bigg(\left(y^1_0 + y^1_1\right) + \frac{3}{5}\left(-y^1_0 + y^1_1 + y^1_2\right)\\
%             &\qquad+ \frac{2}{5}\left(-y^1_0 + y^1_1 - y^1_2\right) + \frac{12}{5} - \fsig{y^1_0 + y^1_1}\\
%             &\qquad- \fsig{-y^1_0 + y^1_1 + y^1_2} - \fsig{-y^1_0 + y^1_1 - y^1_2}\bigg)\Bigg) 
%     \end{aligned}\\
%     &\leq \fmax{2y^1_1 + \frac{1}{5}y^1_2 + \frac{12}{5}}\\
%         &\qquad- \textcolor{purple}{\min\Bigg(2y^1_1 + \frac{1}{5}y^1_2 + \frac{12}{5} -\fsig{y^1_0 + y^1_1}}\\
%         &\qquad\qquad\qquad\textcolor{purple}{- \fsig{-y^1_0 + y^1_1 + y^1_2} - \fsig{-y^1_0 + y^1_1 - y^1_2}\Bigg)}\\
%     &= \fmax{2y^1_1 + \frac{1}{5}y^1_2 + \frac{12}{5}} - \textcolor{purple}{\frac{2}{5}}\\
%     &= \fmax{2\fsig{x^1_1} + \frac{1}{5}\fsig{x^1_2} + \frac{12}{5}} - \textcolor{purple}{\frac{2}{5}}\\
%     &= \textcolor{blue}{\fmax{2\fsig{x^0_0 - x^0_1} + \frac{1}{5}\fsig{x^0_2} + \frac{12}{5}}} - \textcolor{purple}{\frac{2}{5}}\\
%     &= \textcolor{blue}{6\frac{2}{5}} - \textcolor{purple}{\frac{2}{5}} = 6
% \end{align*}
% \guy{TODO (for Guy): work on the equations above}
% \tom{we can use the definition of $E^t$ if we want to make it shorter (but less explicit)}

% The minimized error from the first relaxation is highlighted in
% purple, and the optimization of directly solving the last layer
% instead of minimizing its error is highlighted in blue.  

As we can see, minimizing the errors by using MIP (which is very fast
in practice) allows us to back-substitute bounds with optimal bias,
which yields tighter bounds for the output variable.

\mysubsection{MiniMIP.} While DeepMIP produces very strong bounds, for
each neuron it must solve multiple MIP instances during
back-substitution --- many of them for bounds that may already be
bias-optimal.  This large number of instances to solve can result in a
large overhead, and makes it worthwhile to explore heuristics for only
solving \emph{some} of these instances.

To illustrate this, we propose a particular, aggressive heuristic that
we call \emph{MiniMIP}. Instead of minimizing all error terms during
back-substitution, MiniMIP only solves the final query in this series
--- that is, the query in which the bounds of the current layer are
expressed as sums of \relu{}s of input neurons. This approach
significantly reduces overhead: exactly one MIP instance is solved in
each iteration, regardless of the depth of the layer currently being
processed.  As we later see in our evaluation, even this is already
enough to achieve state-of-the-art performance and very tight bounds;
and the resulting queries can be solved very
efficiently~\cite{TjXiTe17}.

\section{Evaluation}
\label{sec:evaluation}

\mysubsection{Implementation.}  For evaluation purposes, we created a
proof-of-concept implementation of our approach in Python. The
implementation code, alongside all the benchmarks described in this
section, is publicly available online~\cite{tom_zelazny_2022_6982973}.
Our implementation uses the PyTorch library~\cite{pytorch} for computing
the optimal value of $\alpha$ for each \relu{}'s triangle relaxation,
as is done in other tools~\cite{XuZhWaWaJaLiHs20}. We use Gurobi~\cite{Gurobi} as the MIP solver for the minimization of errors and direct concretization of bounds. We ran all
experiments on a compute cluster consisting of Xeon E5-2637 CPUs, and
a 2-hour timeout per experiment.  We note that our implementation
currently runs on CPUs only, and extending it to support GPUs is left
for future work.

\mysubsection{Abstraction refinement cascade.}
For each verification query, 
prior to applying our iterative error minimization scheme, we
configured our implementation to first run a light-weight, 
``ordinary'' symbolic-bound propagation pass. Specifically, we ran a
single pass of the DeepPoly mechanism~\cite{GaGePuVe19}. A similar
technique is applied by other tools~\cite{MuMaSiPuVe22}.

% Similarly to PRIMA,
% we utilize a cascading solution, first attempting to prove the
% properties in question using an implementation of DeepPoly. only
% proceeding to use our approach when the former fails. This also
% applies to our implementation of $\alpha$-CROWN.

\mysubsection{Benchmarks.}
We evaluated our approach on fully-connected, ReLU networks trained
over the MNIST dataset, taken from the ERAN
repository~\cite{eran}. The topologies of the networks we used appear
in Table~\ref{table:networkTopologies}.

\begin{table}[ht]
  \caption{The DNNs used in our evaluation.}
\begin{tabular}{|l|l|l|l|l|l|}
\hline
Dataset                & Model                     & Type                & Neurons & Hidden Layers & Activation            \\ \hline
\multirow{4}{*}{MNIST} & $6\times100$ & \multirow{4}{*}{FC} & 510     & 5             & \multirow{4}{*}{$\relu$} \\ \cline{2-2} \cline{4-5}
                       & $9\times100$ &                     & 810     & 8             &                       \\ \cline{2-2} \cline{4-5}
                       & $6\times200$ &                     & 1010    & 5             &                       \\ \cline{2-2} \cline{4-5}
                       & $9\times200$ &                     & 1610    & 8             &                       \\ \hline
\end{tabular}
\label{table:networkTopologies}
\end{table}

For verification queries, we followed standard
practice~\cite{MuMaSiPuVe22, XuZhWaWaJaLiHs20, KaBaDiJuKo21}, and
attempted to prove the \emph{adversarial robustness} of the first 1000
images of the MNIST test set: that is, we used verification to try and
prove that $\epsilon$-perturbations to correctly classified inputs in the dataset cannot change the classification assigned by the DNN.

We compared the DeepMIP approach (specifically, MiniMIP) to two
state-of-the-art verification approaches~\cite{BaLiJo21}: the PRIMA
solver~\cite{MuMaSiPuVe22}, and our implementation of the
$\alpha$-CROWN method~\cite{XuZhWaWaJaLiHs20}, which represents the
state of the art in symbolic-bound tightening with back-substitution. Indeed, many other verification tools
integrate back-substitution with additional techniques, such as
search-based techniques~\cite{KaHuIbJuLaLiShThWuZeDiKoBa19} or
abstraction-refinement~\cite{AsHaKrMo20}, making it more difficult to
measure the effectiveness of the back-substitution component alone.
However, since the $\alpha$-CROWN implementation in our evaluation also served as the baseline back-substitution method to which we added our methods, any difference between the two is solely due to the addition of our suggested technique. 
The results of our experiments are summarized in
Table~\ref{table:results}.
%\tom{We chose $\alpha$-CROWN because DeepMIP uses it for the
%back-substitution process}
Recall that symbolic-bound propagation techniques are incomplete, and
may fail to prove a given query; the \emph{Solved} columns indicate
the number of instances (out of $1000$) that each method was able to
prove to be robust to adversarial perturbations.  The \emph{Time}
columns indicate the run time of each method (including timeouts), averaged over the $1000$
benchmarks solved.

\begin{table*}[htp]
  \center
  \caption{Comparing DeepMIP to $\alpha$-CROWN and PRIMA.}
\begin{tabular}{|c|c|cc|cc|cc|}
\hline
\multirow{2}{*}{Model} & \multirow{2}{*}{$\epsilon$} &
                                                       \multicolumn{2}{c|}{$\alpha$-CROWN}              & \multicolumn{2}{c|}{PRIMA}                & \multicolumn{2}{c|}{DeepMIP (MiniMIP)}           \\ \cline{3-8} 
                        &                    & \multicolumn{1}{c|}{Solved }      & Time (seconds)    & \multicolumn{1}{c|}{Solved }      & Time (seconds)    & \multicolumn{1}{c|}{Solved }      & Time (seconds)     \\ \hline
$6\times100$            & 0.026              & \multicolumn{1}{c|}{207}        & 38   & \multicolumn{1}{c|}{504}        & 123  & \multicolumn{1}{c|}{581}        & 302   \\ \hline
$9\times100$            & 0.026              & \multicolumn{1}{c|}{223}        & 88   & \multicolumn{1}{c|}{427}        & 252  & \multicolumn{1}{c|}{463}        & 452   \\ \hline
$6\times200$            & 0.015              & \multicolumn{1}{c|}{349}        & 93   & \multicolumn{1}{c|}{652}        & 222  & \multicolumn{1}{c|}{709}        & 801   \\ \hline
$9\times200$            & 0.015              &
                                               \multicolumn{1}{c|}{308}        & 257  & \multicolumn{1}{c|}{600}        & 462  & \multicolumn{1}{c|}{625}        & 1121  \\ \hline

  Total && \multicolumn{1}{c|}{1087} & 476 & \multicolumn{1}{c|}{2183} & 1059 & \multicolumn{1}{c|}{2378} & 2676  \\ \hline
\end{tabular}
\label{table:results}
\end{table*}

Our results clearly indicate the superiority of the bounds discovered
by DeepMIP: indeed, in all categories, our approach was able to solve
the largest number of instances, solving a total of 2378 instances,
compared to 2183 instances solved by PRIMA (198 extra instances solved) and
1087 instances solved by $\alpha$-CROWN (1291 extra instances solved). These
improvements come with an overhead, due to the additional MIP queries
that need to be solved: our approach is approximately 5.6 times slower
than $\alpha$-CROWN, and 2.5 times slower than PRIMA. Furthermore,
DeepMIP timed out on 2 out of the 3829 total benchmarks tested
($\approx 0.05\%$), while PRIMA and $\alpha$-CROWN did not have any timeouts.

The main conclusions that we draw from these experiments are
that
\begin{inparaenum}[(i)]
\item the DeepMIP approach has a significant potential for solving
  queries that other approaches cannot; and
\item additional work, in the form of improved heuristics,
  engineering improvements, and support for GPUs is still required
  to make our approach faster.
\end{inparaenum}
Our results also indicate that a portfolio-based approach, which
starts from light-weight techniques and then progresses towards
DeepMIP for difficult queries, could enjoy the benefits of both
worlds.

\section{Related Work}
\label{sec:relatedWork}
 
The topic of DNN verification has been receiving significant attention
from the formal methods community, and various tools and methods and
have been proposed for addressing it.  These include techniques that
leverage SMT solvers (e.g.,~\cite{HuKwWaWu17, PuTa10,
  KaHuIbJuLaLiShThWuZeDiKoBa19, WuZeKaBa22}), LP and MILP solvers
(e.g.,~\cite{LoMa17, Eh17, TjXiTe17, BuTuToKoMu18}), reachability
analysis~\cite{SuKhSh19}, abstraction-refinement
techniques~\cite{ElGoKa20, AsHaKrMo20, ElCoKa22}, and many others. The
techniques most related to DeepMIP are those that rely on the
propagation of symbolic bounds using abstract interpretation
(e.g.,~\cite{WaPeWhYaJa18, GeMiDrTsChVe18, WeZhChSoHsBoDhDa18,
  TrBkJo20}). Recent work has also extended beyond answering binary
questions about DNNs, instead targeting tasks such as automated DNN
repair~\cite{KoLoJaBl20, GoAdKeKa20}, DNN
simplification~\cite{GoFeMaBaKa20,LaKa21}, ensemble
selection~\cite{AmKaSc22}, and quantitative
verification and optimization~\cite{BaShShMeSa19, StWuZeJuKaBaKo21};
and also the verification of recurrent neural
networks~\cite{ZhShGuGuLeNa20, JaBaKa20, RyChBaSiDaVe21} and
reinforcement-learning based systems~\cite{JiTiZhWeZh22, ElKaKaSc21,
  AmScKa21}. Our proposed techniques could be integrated into any
number of these approaches.

Bound propagation has been playing a significant part in DNN
verification efforts for the past few years. Starting with
interval-arithmetic-based propagation~\cite{KaBaDiJuKo21} and
optimization queries for individual neurons~\cite{Eh17, TjXiTe17},
these approaches have progressed to use various relaxations and
over-approximations for individual neurons~\cite{GaGePuVe19,
  GeMiDrTsChVe18, WaPeWhYaJa18} and sets thereof~\cite{MuMaSiPuVe22,
  SiGaPuVe19, OsBaKa22}, culminating in highly sophisticated
approaches~\cite{MuMaSiPuVe22, XuZhWaWaJaLiHs20}. We consider our work
as another step in this very promising research direction.

\section{Conclusion and Future Work}\label{sec:conclusion}
We presented an enhancement to the popular back-substitution
procedure, which includes a formulation of the over-approximation
errors introduced during back-substitution. These errors can then be
minimized, in order to greatly tighten the resulting bounds. Our
approach achieves tighter bounds than state-of-the-art approaches, but
at the cost of longer running times; and we are currently exploring
methods for expediting it. Specifically, moving forward, we intend to
focus on adding support for GPUs; on better refinement heuristics; on
better MIP encoding~\cite{AnHuTjVi18}; and also on improving the core
algorithm to utilize previously calculated bounds and errors.
Furthermore, we intend to generalize our methods to other abstract
domains, and also to integrate them with search-based techniques.

\section*{Acknowledgements}
The project was partially supported by the Israel Science Foundation
(grant number 683/18) and by the Binational Science Foundation (grant
number 2020250).

\newpage

{
% \bibliographystyle{abbrv}
% \bibliography{bibliography}

\begin{thebibliography}{10}

\bibitem{AkKeLoPi19}
M.~Akintunde, A.~Kevorchian, A.~Lomuscio, and E.~Pirovano.
\newblock {Verification of RNN-Based Neural Agent-Environment Systems}.
\newblock In {\em Proc. 33rd AAAI Conf. on Artificial Intelligence (AAAI)},
  pages 197--210, 2019.

\bibitem{Al19}
M.~AlQuraishi.
\newblock {AlphaFold at CASP13}.
\newblock {\em Bioinformatics}, 35(22):4862--4865, 2019.

\bibitem{AmKaSc22}
G.~Amir, G.~Katz, and M.~Schapira.
\newblock {Verification-Aided Deep Ensemble Selection}.
\newblock In {\em Proc. 22nd Int. Conf. on Formal Methods in Computer-Aided
  Design (FMCAD)}, 2022.

\bibitem{AmScKa21}
G.~Amir, M.~Schapira, and G.~Katz.
\newblock {Towards Scalable Verification of Deep Reinforcement Learning}.
\newblock In {\em Proc. 21st Int. Conf. on Formal Methods in Computer-Aided
  Design (FMCAD)}, pages 193--203, 2021.

\bibitem{AmOlStChScMa16}
D.~Amodei, C.~Olah, J.~Steinhardt, P.~Christiano, J.~Schulman, and D.~Man\'e.
\newblock {Concrete Problems in AI Safety}, 2016.
\newblock Technical Report. \url{https://arxiv.org/abs/1606.06565}.

\bibitem{AnHuTjVi18}
R.~Anderson, J.~Huchette, C.~Tjandraatmadja, and J.~Vielma.
\newblock {Strong Mixed-Integer Programming Formulations for Trained Neural
  Networks}, 2018.
\newblock Technical Report. \url{http://arxiv.org/abs/1811.08359}.

\bibitem{AsHaKrMo20}
P.~Ashok, V.~Hashemi, J.~Kretinsky, and S.~Mohr.
\newblock {DeepAbstract: Neural Network Abstraction for Accelerating
  Verification}.
\newblock In {\em Proc. 18th Int. Symp. on Automated Technology for
  Verification and Analysis (ATVA)}, pages 92--107, 2020.

\bibitem{AvBlChHeKoPr19}
G.~Avni, R.~Bloem, K.~Chatterjee, T.~Henzinger, B.~Konighofer, and S.~Pranger.
\newblock {Run-Time Optimization for Learned Controllers through Quantitative
  Games}.
\newblock In {\em Proc. 31st Int. Conf. on Computer Aided Verification (CAV)},
  pages 630--649, 2019.

\bibitem{BaLiJo21}
S.~Bak, C.~Liu, and T.~Johnson.
\newblock {The Second International Verification of Neural Networks Competition
  (VNN-COMP 2021): Summary and Results}, 2021.
\newblock Technical Report. \url{http://arxiv.org/abs/2109.00498}.

\bibitem{BaShShMeSa19}
T.~Baluta, S.~Shen, S.~Shinde, K.~Meel, and P.~Saxena.
\newblock {Quantitative Verification of Neural Networks And its Security
  Applications}.
\newblock In {\em Proc. 26th ACM Conf. on Computer and Communication Security
  (CCS)}, 2019.

\bibitem{BaIoLaVyNoCr16}
O.~Bastani, Y.~Ioannou, L.~Lampropoulos, D.~Vytiniotis, A.~Nori, and
  A.~Criminisi.
\newblock {Measuring Neural Net Robustness with Constraints}.
\newblock In {\em Proc. 30th Conf. on Neural Information Processing Systems
  (NIPS)}, 2016.

\bibitem{BoDeDwFiFlGoJaMoMuZhZhZhZi16}
M.~Bojarski, D.~Del~Testa, D.~Dworakowski, B.~Firner, B.~Flepp, P.~Goyal,
  L.~Jackel, M.~Monfort, U.~Muller, J.~Zhang, X.~Zhang, J.~Zhao, and K.~Zieba.
\newblock {End to End Learning for Self-Driving Cars}, 2016.
\newblock Technical Report. \url{http://arxiv.org/abs/1604.07316}.

\bibitem{BuTuToKoMu18}
R.~Bunel, I.~Turkaslan, P.~Torr, P.~Kohli, and P.~Mudigonda.
\newblock {A Unified View of Piecewise Linear Neural Network Verification}.
\newblock In {\em Proc. 32nd Conf. on Neural Information Processing Systems
  (NeurIPS)}, pages 4795--4804, 2018.

\bibitem{DrFrGhKiRaVaSe19}
T.~Dreossi, D.~Fremont, S.~Ghosh, E.~Kim, H.~Ravanbakhsh, M.~Vazquez-Chanlatte,
  and S.~Seshia.
\newblock {VerifAI: A Toolkit for the Formal Design and Analysis of Artificial
  Intelligence-Based Systems}.
\newblock In {\em Proc. 31st Int. Conf. on Computer Aided Verification (CAV)},
  pages 432--442, 2019.

\bibitem{Eh17}
R.~Ehlers.
\newblock {Formal Verification of Piece-Wise Linear Feed-Forward Neural
  Networks}.
\newblock In {\em Proc. 15th Int. Symp. on Automated Technology for
  Verification and Analysis (ATVA)}, pages 269--286, 2017.

\bibitem{ElCoKa22}
Y.~Elboher, E.~Cohen, and G.~Katz.
\newblock {Neural Network Verification using Residual Reasoning}.
\newblock In {\em Proc. 20th Int. Conf. on Software Engineering and Formal
  Methods (SEFM)}, 2022.

\bibitem{ElGoKa20}
Y.~Elboher, J.~Gottschlich, and G.~Katz.
\newblock {An Abstraction-Based Framework for Neural Network Verification}.
\newblock In {\em Proc. 32nd Int. Conf. on Computer Aided Verification (CAV)},
  pages 43--65, 2020.

\bibitem{ElKaKaSc21}
T.~Eliyahu, Y.~Kazak, G.~Katz, and M.~Schapira.
\newblock {Verifying Learning-Augmented Systems}.
\newblock In {\em Proc. Conf. of the ACM Special Interest Group on Data
  Communication on the Applications, Technologies, Architectures, and Protocols
  for Computer Communication (SIGCOMM)}, pages 305--318, 2021.

\bibitem{eran}
ERAN.
\newblock {The ERAN Repository}, 2022.
\newblock \url{https://github.com/eth-sri/eran}.

\bibitem{EyEvFeLiRaXiPrKoSo18}
K.~Eykholt, I.~Evtimov, E.~Fernandes, B.~Li, A.~Rahmati, C.~Xiao, A.~Prakash,
  T.~Kohno, and D.~Song.
\newblock {Robust Physical-World Attacks on Deep Learning Visual
  Classification}.
\newblock In {\em Proc. IEEE Conf. on Computer Vision and Pattern Recognition
  (CVPR)}, pages 1625--1634, 2018.

\bibitem{GeMiDrTsChVe18}
T.~Gehr, M.~Mirman, D.~Drachsler-Cohen, E.~Tsankov, S.~Chaudhuri, and
  M.~Vechev.
\newblock {AI2: Safety and Robustness Certification of Neural Networks with
  Abstract Interpretation}.
\newblock In {\em Proc. 39th IEEE Symposium on Security and Privacy (S\&P)},
  2018.

\bibitem{GoFeMaBaKa20}
S.~Gokulanathan, A.~Feldsher, A.~Malca, C.~Barrett, and G.~Katz.
\newblock {Simplifying Neural Networks using Formal Verification}.
\newblock In {\em Proc. 12th NASA Formal Methods Symposium (NFM)}, pages
  85--93, 2020.

\bibitem{GoAdKeKa20}
B.~Goldberger, Y.~Adi, J.~Keshet, and G.~Katz.
\newblock {Minimal Modifications of Deep Neural Networks using Verification}.
\newblock In {\em Proc. 23rd Int. Conf. on Logic for Programming, Artificial
  Intelligence and Reasoning (LPAR)}, pages 260--278, 2020.

\bibitem{FoBeCu16}
I.~Goodfellow, Y.~Bengio, and A.~Courville.
\newblock {\em {Deep Learning}}.
\newblock MIT Press, 2016.

\bibitem{Gu17}
D.~Gunning.
\newblock {Explainable Artificial Intelligence (XAI)}, 2017.
\newblock Defense Advanced Research Projects Agency (DARPA) Project.

\bibitem{Gurobi}
Gurobi.
\newblock {The Gurobi MILP Solver}, 2021.
\newblock \url{https://www.gurobi.com/}.

\bibitem{HuKwWaWu17}
X.~Huang, M.~Kwiatkowska, S.~Wang, and M.~Wu.
\newblock {Safety Verification of Deep Neural Networks}.
\newblock In {\em Proc. 29th Int. Conf. on Computer Aided Verification (CAV)},
  pages 3--29, 2017.

\bibitem{JaBaKa20}
Y.~Jacoby, C.~Barrett, and G.~Katz.
\newblock {Verifying Recurrent Neural Networks using Invariant Inference}.
\newblock In {\em Proc. 18th Int. Symposium on Automated Technology for
  Verification and Analysis (ATVA)}, pages 57--74, 2020.

\bibitem{JiTiZhWeZh22}
P.~Jin, J.~Tian, D.~Zhi, X.~Wen, and M.~Zhang.
\newblock {Trainify: A CEGAR-Driven Training and Verification Framework for
  Safe Deep Reinforcement Learning}.
\newblock In {\em Proc. 34th Int. Conf. on Computer Aided Verification (CAV)},
  pages 193--218, 2022.

\bibitem{JuLoBrOwKo16}
K.~Julian, J.~Lopez, J.~Brush, M.~Owen, and M.~Kochenderfer.
\newblock {Policy Compression for Aircraft Collision Avoidance Systems}.
\newblock In {\em Proc. 35th Digital Avionics Systems Conf. (DASC)}, pages
  1--10, 2016.

\bibitem{KaBaDiJuKo21}
G.~Katz, C.~Barrett, D.~Dill, K.~Julian, and M.~Kochenderfer.
\newblock {Reluplex: a Calculus for Reasoning about Deep Neural Networks}.
\newblock {\em Formal Methods in System Design (FMSD)}, 2021.

\bibitem{KaHuIbJuLaLiShThWuZeDiKoBa19}
G.~Katz, D.~Huang, D.~Ibeling, K.~Julian, C.~Lazarus, R.~Lim, P.~Shah,
  S.~Thakoor, H.~Wu, A.~Zelji\'c, D.~Dill, M.~Kochenderfer, and C.~Barrett.
\newblock {The Marabou Framework for Verification and Analysis of Deep Neural
  Networks}.
\newblock In {\em Proc. 31st Int. Conf. on Computer Aided Verification (CAV)},
  pages 443--452, 2019.

\bibitem{KoKoKiAtAs20}
W.~Kokke, E.~Komendantskaya, D.~Kienitz, R.~Atkey, and D.~Aspinall.
\newblock {Neural Networks, Secure by Construction: An Exploration of
  Refinement Types}.
\newblock In {\em Proc. 18th Asian Symposium on Programming Languages and
  Systems (APLAS)}, pages 67--85, 2020.

\bibitem{KoLoJaBl20}
B.~K\"{o}nighofer, F.~Lorber, N.~Jansen, and R.~Bloem.
\newblock {Shield Synthesis for Reinforcement Learning}.
\newblock In {\em Proc. Int. Symposium On Leveraging Applications of Formal
  Methods, Verification and Validation (ISoLA)}, pages 290--306, 2020.

\bibitem{LaKa21}
O.~Lahav and G.~Katz.
\newblock {Pruning and Slicing Neural Networks using Formal Verification}.
\newblock In {\em Proc. 21st Int. Conf. on Formal Methods in Computer-Aided
  Design (FMCAD)}, pages 183--192, 2021.

\bibitem{LoMa17}
A.~Lomuscio and L.~Maganti.
\newblock {An Approach to Reachability Analysis for Feed-Forward ReLU Neural
  Networks}, 2017.
\newblock Technical Report. \url{http://arxiv.org/abs/1706.07351}.

\bibitem{MuMaSiPuVe22}
M.~M\"{u}ller, G.~Makarchuk, G.~Singh, M.~Puschel, and M.~Vechev.
\newblock {PRIMA: General and Precise Neural Network Certification via Scalable
  Convex Hull Approximations}.
\newblock In {\em Proc. 49th ACM SIGPLAN Symposium on Principles of Programming
  Languages (POPL)}, 2022.

\bibitem{OsBaKa22}
M.~Ostrovsky, C.~Barrett, and G.~Katz.
\newblock {An Abstraction-Refinement Approach to Verifying Convolutional Neural
  Networks}.
\newblock In {\em Proc. 20th. Int. Symposium on Automated Technology for
  Verification and Analysis (ATVA)}, 2022.

\bibitem{PuTa10}
L.~Pulina and A.~Tacchella.
\newblock {An Abstraction-Refinement Approach to Verification of Artificial
  Neural Networks}.
\newblock In {\em Proc. 22nd Int. Conf. on Computer Aided Verification (CAV)},
  pages 243--257, 2010.

\bibitem{pytorch}
PyTorch.
\newblock {The PyTorch Library}, 2022.
\newblock \url{https://pytorch.org/}.

\bibitem{RyChBaSiDaVe21}
W.~Ryou, J.~Chen, M.~Balunovic, G.~Singh, A.~Dan, and M.~Vechev.
\newblock {Scalable Polyhedral Verification of Recurrent Neural Networks}.
\newblock In {\em Proc. 33rdd Int. Conf. on Computer Aided Verification (CAV)},
  pages 225--248, 2021.

\bibitem{SiHuMaGuSiVaScAnPaLaDi16}
D.~Silver, A.~Huang, C.~Maddison, A.~Guez, L.~Sifre, G.~Van Den~Driessche,
  J.~Schrittwieser, I.~Antonoglou, V.~Panneershelvam, M.~Lanctot, and
  S.~Dieleman.
\newblock {Mastering the Game of Go with Deep Neural Networks and Tree Search}.
\newblock {\em Nature}, 529(7587):484--489, 2016.

\bibitem{SiZi14}
K.~Simonyan and A.~Zisserman.
\newblock {Very Deep Convolutional Networks for Large-Scale Image Recognition},
  2014.
\newblock Technical Report. \url{http://arxiv.org/abs/1409.1556}.

\bibitem{SiGaPuVe19}
G.~Singh, R.~Ganvir, M.~Puschel, and M.~Vechev.
\newblock {Beyond the Single Neuron Convex Barrier for Neural Network
  Certification}.
\newblock In {\em Proc. 33rd Conf. on Neural Information Processing Systems
  (NeurIPS)}, 2019.

\bibitem{GaGePuVe19}
G.~Singh, T.~Gehr, M.~Puschel, and M.~Vechev.
\newblock {An Abstract Domain for Certifying Neural Networks}.
\newblock In {\em Proc. 46th ACM SIGPLAN Symposium on Principles of Programming
  Languages (POPL)}, 2019.

\bibitem{StWuZeJuKaBaKo21}
C.~Strong, H.~Wu, A.~Zelji\'c, K.~Julian, G.~Katz, C.~Barrett, and
  M.~Kochenderfer.
\newblock {Global Optimization of Objective Functions Represented by ReLU
  Networks}.
\newblock {\em Journal of Machine Learning}, pages 1--28, 2021.

\bibitem{SuKhSh19}
X.~Sun, K.~H., and Y.~Shoukry.
\newblock {Formal Verification of Neural Network Controlled Autonomous
  Systems}.
\newblock In {\em Proc. 22nd ACM Int. Conf. on Hybrid Systems: Computation and
  Control (HSCC)}, 2019.

\bibitem{SzZaSuBrErGoFe13}
C.~Szegedy, W.~Zaremba, I.~Sutskever, J.~Bruna, D.~Erhan, I.~Goodfellow, and
  R.~Fergus.
\newblock {Intriguing Properties of Neural Networks}, 2013.
\newblock Technical Report. \url{http://arxiv.org/abs/1312.6199}.

\bibitem{TjXiTe17}
V.~Tjeng, K.~Xiao, and R.~Tedrake.
\newblock {Evaluating Robustness of Neural Networks with Mixed Integer
  Programming}, 2017.
\newblock Technical Report. \url{http://arxiv.org/abs/1711.07356}.

\bibitem{TrBkJo20}
H.~Tran, S.~Bak, and T.~Johnson.
\newblock {Verification of Deep Convolutional Neural Networks Using
  ImageStars}.
\newblock In {\em Proc. 32nd Int. Conf. on Computer Aided Verification (CAV)},
  pages 18--42, 2020.

\bibitem{WaPeWhYaJa18}
S.~Wang, K.~Pei, J.~Whitehouse, J.~Yang, and S.~Jana.
\newblock {Formal Security Analysis of Neural Networks using Symbolic
  Intervals}.
\newblock In {\em Proc. 27th USENIX Security Symposium}, 2018.

\bibitem{WeZhChSoHsBoDhDa18}
T.-W. Weng, H.~Zhang, H.~Chen, Z.~Song, C.-J. Hsieh, D.~Boning, I.~Dhillon, and
  L.~Daniel.
\newblock {Towards Fast Computation of Certified Robustness for ReLU Networks},
  2018.
\newblock Technical Report. \url{http://arxiv.org/abs/1804.09699}.

\bibitem{WuZeKaBa22}
H.~Wu, A.~Zelji\'c, G.~Katz, and C.~Barrett.
\newblock {Efficient Neural Network Analysis with Sum-of-Infeasibilities}.
\newblock In {\em Proc. 28th Int. Conf. on Tools and Algorithms for the
  Construction and Analysis of Systems (TACAS)}, pages 143--163, 2022.

\bibitem{XuZhWaWaJaLiHs20}
K.~Xu, H.~Zhang, S.~Wang, Y.~Wang, S.~Jana, X.~Lin, and C.-J. Hsieh.
\newblock {Fast and Complete: Enabling Complete Neural Network Verification
  with Rapid and Massively Parallel Incomplete Verifiers}, 2020.
\newblock Technical Report. \url{http://arxiv.org/abs/2011.13824}.

\bibitem{tom_zelazny_2022_6982973}
T.~Zelazny, H.~Wu, C.~Barrett, and G.~Katz.
\newblock {DeepMIP Code}, 2022.
\newblock \url{https://doi.org/10.5281/zenodo.6982973}.

\bibitem{ZhShGuGuLeNa20}
H.~Zhang, M.~Shinn, A.~Gupta, A.~Gurfinkel, N.~Le, and N.~Narodytska.
\newblock {Verification of Recurrent Neural Networks for Cognitive Tasks via
  Reachability Analysis}.
\newblock In {\em Proc. 24th European Conf. on Artificial Intelligence (ECAI)},
  pages 1690--1697, 2020.

\end{thebibliography}

}

\newpage
\onecolumn

\begin{appendices}
\section{Relaxation Matrices}

\label{sec:appendix:relaxationMatrices}

The matrices $R^t_U$ and $R^t_L$ are how we apply the triangle
relaxation during back-substitution over layer $t$. for example if:
\[
  x^{i+1}_j = \fsig{x^i_0} - 2\fsig{x^i_1}
\]
then in order to find a linear upper bound for $x^{i+1}_j$,  we need
to replace $\fsig{x^i_0}$ with its triangle-relaxation upper bound
(since it has a positive coefficient), and $\fsig{x^i_1}$ with its
triangle-relaxation lower bound. This gives rise to:
\[
  x^{i+1}_j \leq \frac{u^i_0}{u^i_0 - l^i_0} (x^i_0 - l^i_0) - 2\alpha
  x^i_1
\]
which can be written as (for some constant term $d$): 
\[ x^{i+1}_j \leq \frac{u^i_0}{u^i_0 - l^i_0} x^i_0 - 2\alpha x^i_1 + d
\]
Written as a vector product:
\[
  x^{i+1}_j =
  \begin{bmatrix}
    1 & -2\\
  \end{bmatrix}
  \cdot
  \begin{bmatrix}
    \fsig{x^i_0}\\ \fsig{x^i_1}\\
  \end{bmatrix}
  \leq
  \begin{bmatrix}
    1 & -2\\
  \end{bmatrix}
  \cdot
  \begin{bmatrix}
    \frac{u^i_0}{u^i_0 - l^i_0} & 0\\ 0 & \alpha\\
  \end{bmatrix}
  \cdot
  \begin{bmatrix}
    x^i_0\\ x^i_1\\
  \end{bmatrix}
  + d
\]
We use $R^i_U$ to denote the matrix that was used to relax
$\fsig{x^i}$, and observe that it depends on the weights/coefficients
of each non-linearity about to be relaxed, and also on the existence of $[l^i, u^i]$ in order to compute the corresponding relaxations. Formally we define the matrix $R^t_U(\omega^t, l^t, u^t)$ as:
\[ R^t_U(\omega^t, l^t, u^t)[i,j] = 0 \qquad i \neq j\]
\[ R^t_U(\omega^t, l^t, u^t)[i,i] \equiv  \begin{cases} 
    1  & \text{if\ } l^t_i \geq 0\\
    0  & \text{if\ } u^t_i \leq 0\\
    \frac{u^t_i}{u^t_i - l^t_i} & \text{if\ }\omega^t_i \geq 0 \text{\ and\ } l^t_i \leq 0 \leq u^t_i\\
    \alpha  & \text{if\ } \omega^t_i \leq 0 \text{\ and\ } l^t_i \leq 0 \leq u^t_i\\
  \end{cases}\] where $\omega^t$ is a row vector such that
$\omega^t_i$ contains the coefficient of $\fsig{x^t_i}$, and
$l^t, u^t$ are vectors such that $l^t_i \leq x^t_i \leq
u^t_i$. Similarly, we
define $R^t_L(\omega^t, l^t, u^t)$ as:
\[ R^t_L(\omega^t, l^t, u^t)[i,j] = 0 \qquad i \neq j\]
\[ R^t_L(\omega^t, l^t, u^t)[i,i] \equiv  \begin{cases} 
    1  & \text{if\ } l^t_i \geq 0\\
    0  & \text{if\ } u^t_i \leq 0\\
    \frac{u^t_i}{u^t_i - l^t_i} & \text{if\ }\omega^t_i \leq 0 \text{\ and\ } l^t_i \leq 0 \leq u^t_i\\
    \alpha  & \text{if\ } \omega^t_i \geq 0 \text{\ and\ } l^t_i \leq 0 \leq u^t_i\\
\end{cases}\]

We note that there exists similar matrices for updating the constant term
during back-substitution; we omit them to reduce clutter.
Furthermore, when it is clear from context, we write $R^t_L, R^t_U$ instead of $R^t_L(\omega^t,l^t,u^t), R^t_U(\omega^t,l^t,u^t)$.

\newpage

  \section{The DeepMIP Algorithm}
\label{sec:appendix:pseudocode}

Below is the pseudocode for the DeepMIP algorithm:

\algnewcommand{\LineComment}[1]{\State \(\triangleright\) #1}

\begin{algorithm}
	\caption{The DeepMIP algorithm}
	\hspace*{\algorithmicindent} \textbf{Input:} DNN $N$, input constraints $\inputDomain$, output constraints $\outputDomain$ \\
	\hspace*{\algorithmicindent} \textbf{Output:} bounds $B$ for every neuron in $N$
	\begin{algorithmic}[1]
	    \State $\Call{InitializeInputConstraints}{N, \inputDomain}$
		\For{$i=1 \ldots k$} \label{line:iterateLayers}
    		\For{$j=0 \ldots n_i$} \label{line:iterateNeurons}
    		    \State $\Call{InitializeAlphas}{N,i,j,\text{LB}}$
    		    \State $N^i_j[\text{concreteLB}] \gets \Call{BackSubstitute}{N, i, j,\text{LB}}$
    		    \State $\Call{InitializeAlphas}{N,i,j,\text{UB}}$
    		    \State $N^i_j[\text{concreteUB}] \gets \Call{BackSubstitute}{N, i, j, \text{UB}}$
    		    \State $\Call{ApplyActivation}{N, i, j}$
        % 		\State $v^j_{i_1},\ldots,v^j_{i_k} \gets N(x_j)$  \Comment{Compute the separation layers' assignments} 
    		\EndFor \label{line:iterateLayers}
		\EndFor \label{line:iterateNeurons}
		\State \Return{$\Call{TestOutputConstraints}{N, \outputDomain}$}
	\end{algorithmic}
	\label{alg:solution}
\end{algorithm}
\begin{algorithm}
	\caption{BackSubstitute}
	\hspace*{\algorithmicindent} \textbf{Input:} DNN $N$, $i$ layer index, $j$ neuron index, $\text{boundType}$ the requested bound type \\
	\hspace*{\algorithmicindent} \textbf{Output:} concrete bound for (either lower or upper depending on $\text{boundType}$ for the requested neuron
	\begin{algorithmic}[1]
	   % \Comment{first we initialize the bound depending on if we are calculating a lower or upper bound}
	    \If{$\text{boundType} = \text{LB}$}
	        \State $\text{bound} \gets -\infty$
	       % \State $\text{product} \gets \text{LProduct}$
	    \ElsIf{$\text{boundType} = \text{UB}$}
	        \State $\text{bound} \gets \infty$
	       % \State $\text{product} \gets \text{UProduct}$
	    \EndIf
	    \State $p \gets i-1$ \Comment{p is the index of the layer containing the variables of the symbolic bound}
	    \State $s \gets W^p_j$ \Comment{s is a vector containing the coefficients of the symbolic bound}
	    \State $b \gets b^p_j$ \Comment{b is the constant term of
              the symbolic bound}
            \LineComment{the bound is $\boldsymbol{s} \cdot
              \boldsymbol{y^p} + b = \boldsymbol{s} \cdot
              \boldsymbol{\fsig{x^p}} + b = \boldsymbol{s} \cdot
              \boldsymbol{\fsig{W^{p-1}y^{p-1}}} + b$; and we update $s$ and $b$ to maintain the validity of the bound during back-substitution, using the triangle relaxation}
		\While{$p\geq1$} \label{line:backsubLoop}
	        \State $\text{candidateBound} \gets \Call{ConcretizePartialMIP}{s,b,N,p,\text{boundType}}$ \Comment{concretize at each layer using MIP}
	        \State $\text{bound} \gets \Call{PickTighterBound}{\text{candidateBound}, \text{bound}}$ \Comment{keep the best bound found so far}
	        \State $b = b - \Call{boundError}{s,b,p,\text{boundType}}$ \Comment{improve detached bounds}
		    \State $s,b \gets \Call{SubstituteWithPreviousLayer}{s,b,p}$ \Comment{back-substitute a single layer back using triangle relaxation}
		    \State $p \gets p-1$
		\EndWhile \label{line:backsubLoop}
		\Return{$bound$}
	\end{algorithmic}
	\label{alg:solution}
\end{algorithm}

Where the methods mentioned in the pseudocode are:
\begin{itemize}
\item \emph{InitializeAlphas}:
A method that
can implement any heuristics that chooses the $\alpha$ parameters for
the lower bound of the triangle approximation of a \relu{}.
\item \emph{ApplyActivation}: a method that propagates
  the concrete bounds through the activation layer.
\item \emph{ConcretizePartialMIP}: a method that obtains a concrete
  bound using the current symbolic bound.
\item \emph{BoundError}: a method that obtains a concrete bound on the
  error.
  \end{itemize}

In the paper we use a MIP solver for both \emph{ConcretizePartialMIP} and \emph{BoundError}; however, any method that can obtain a non-trivial bound could be used instead.

\end{appendices}
\end{document}